BACHELOR OF SCIENCE DEGREE IN ELECTRICAL ENGINEERING

Senior Design Project Report

Department of Electrical Engineering
Qatar University

# Design, Development and Evaluation of a UAV to Study Air Quality in Qatar

Report by
Khalid Al-Hajjaji
Mouadh Ezzin
Husain Khamdan
Abdelhakim El Hassani

Supervisor
Dr. Nizar Zorba

Date
25/05/2017

# DECLARATION STATEMENT

We, the undersigned students, confirm that the work submitted in this project report is entirely our own and has not been copied from any other source. Any material that has been used from other sources has been properly cited and acknowledged in the report.

We are fully aware that any copying or improper citation of references/sources used in this report will be considered plagiarism, which is a clear violation of the Code of Ethics of Qatar University.

In addition, we have read and understood the legal consequences of committing any violation of the Qatar University's Code of Ethics.

| | Student Name | Student ID | Signature | Date |
|---|---|---|---|---|
| 1 | Khalid Al-Hajjaji | 201206570 | | 22/12/2016 |
| 2 | Mouadh Ezzin | 201203100 | | 22/12/2016 |
| 3 | Husain Khamdan | 201205752 | | 22/12/2016 |
| 4 | Abdelhakim El Hassani | 20098481 | | 22/12/2016 |

# ABSTRACT

Measuring gases for air quality monitoring is a challenging task that claims a lot of time of observation and large numbers of sensors. The aim of this project is to develop a partially autonomous unmanned aerial vehicle (UAV) equipped with sensors, in order to monitor and collect air quality real time data in designated areas and send it to the ground base. This project is designed and implemented by a multidisciplinary team from electrical and computer engineering departments. The electrical engineering team responsible for implementing air quality sensors for detecting real time data and transmit it from the plane to the ground. On the other hand, the computer engineering team is in charge of Interface sensors and provide platform to view and visualize air quality data and live video streaming. The proposed project contains several sensors to measure Temperature, Humidity, Dust, CO, $CO_2$ and $O_3$. The collected data is transmitted to a server over a wireless internet connection and the server will store, and supply these data to any party who has permission to access it through android phone or website in semi-real time. The developed UAV has carried several field tests in Al Shamal airport in Qatar, with interesting results and proof of concept outcomes.

# ACKNOWLEDGEMENTS

The whole team would like to thank BOEING for their generous funding for this project and the provided support to it. Electrical Engineering team would like to thank Dr. Nizar Zorba the supervisor of the project for his support and follow on every step during the senior design project. Furthermore, the team is very thankful to Dr. Ryan Riley from Computer Engineering Department and their team for their contentious support in this project. In addition, the group members appreciate the help of Prof. Nasr Bensalah form chemistry department and Dr. Farid Touati from Electrical Engineering Department. We extend our gratitude to the Computer Engineering team for their assistance and support to successfully complete our project. Most of all, the team would like to thank our colleague Husain Alabbad and all members of Qatar RC Sport center for their contribution to the success of this project.

# TABLE OF CONTENTS

# LIST OF FIGURES

# LIST OF TABLES

# GLOSSARY

## List of Abbreviations

| | |
|---|---|
| BEC | Battery Eliminator Circuit |
| CO | Carbon monoxide |
| $CO_2$ | Carbon dioxide |
| CPU | Central Processing Unit |
| DC | Direct Current |
| ESC | Electronic Speed Control |
| GPS | Global Position System |
| IDE | Integrated Development Environment |
| IoT | Internet of Things |
| Li-Po | Lithium-ion Polymer |
| LPG | Liquefied Petroleum Gas |
| NMHCs | Non-Methane Hydrocarbons |
| $NO_2$ | Nitrogen dioxide |
| PC | Personal Computer |
| PV | Photovoltaic |
| QEERI | Qatar Environment and Energy Research Institute |
| RAM | Random Access Memory |
| RC | Remote Control |
| RPM | Revolutions Per Minute |
| STC | Standard Test Conditions |
| UAV | Unmanned Aerial Vehicle |
| USB | Universal Serial Bus |

# List of Symbols

| | |
|---|---|
| a | Acceleration |
| α | Angle of attack |
| $A$ | Wing area |
| $C_d$ | Drag coefficient |
| $C_l$ | Lift coefficient |
| d | Propeller diameter |
| $F_w$ | Weighted force |
| $F_l$ | Lift force |
| $F_r$ | Friction force |
| $F_T$ | Thrust force |
| $F_d$ | Drag force |
| g | Gravity force |
| Kv | RPM constant of a motor |
| m | Mass |
| $P_{elec}$ | Electrical power |
| $P_{mech}$ | Mechanical power |
| $\rho air$ | Density of the air in kg/m3 |
| Re | Reynold number |
| µ | Rolling friction |
| V | Velocity |
| ɣ | Kinematic viscosity |
| η | Motor efficiency |

# Chapter 1 INTRODUCTION

The introduction will describe the background and problem definition of using Unmanned Aerial Vehicle (UAV) for monitoring air quality to gather data. Furthermore, it will define the objectives and project plan.

## *1.1 Background*

UAVs have been developing since World War I; the Curtiss N2C-2 drone was developed in 1937 by USA navy as the first radio-controlled airplane. In 1941, Radio plane OQ-2 was developed by Reginald Denny as the first large-scale produced UAV to be used in military purposes. At their start, UAVs were highly classified as military technology with no widespread civilian usage. As technology advanced and with the emerging of microcontrollers and micro-computers as well as lifting of restrictions on flying hobbyist’s UAVs in most countries, UAVs became common, thus, civilian low cost project that rely on UAVs became economically and logistically accessible. Today, UAV’s are being developed to deliver shipments, take aerial footage of sport matches, as well as many other applications [1].

## *1.2 Problem Definition*

Measuring air quality is important to make sure that the general public, governmental agencies and any involved party is conscious of the state of pollution, and to trigger taking the required precautions to ensure the safety of the population. As indicated by a report from the world health organization (WHO), around 7 million individuals die per year from causes related to air contamination. The aims of this project is to build a system that helps environmental researchers and other parties interested in monitoring air quality. As well as to provide those researchers with the necessary tools to visualize and analyze the gathered data in user-friendly interfaces. The system will gather and transmit real time air quality data and provide live streaming to the control center. However, a transmission delay could be an issue especially when it’s on auto pilot mood. In addition, component selectivity and flying time are another problem in system design.

Based on WHO, the most gases that have significant impact on people and environment are listed in Table 1-1 with their sources and impacts on air quality [2].

**Table 1-1: Source and impact of air quality monitoring gases**

| Parameter | Source | Impact |
|---|---|---|
| **CO** | - Vehicular exhaust.<br>- Product by incomplete burning of fuel. | The ability of the bloods to carry oxygen will stop. Exposure for small amounts causes dizziness, headaches and slowed reaction times. Exposure for Large amounts are deadly. |
| **$CO_2$** | - Combustion and motor vehicle.<br>- Cement production.<br>- Industry.<br>- Respiration of animals and people. | Greenhouse gas it causes trap heat and $CO_2$ can create an oxygen deficiency and it is the main danger for in asphyxiation. |
| **NOx** | - High temperature heat source.<br>- Burning fossil fuels.<br>- In home, gas cookers and cigarette smoke | Exposed to higher levels of $NO_2$ Cause a problem for eyes, nose, throat and respiratory problems especially in asthmatics. |
| **SOx** | - Volcanoes.<br>- Various industrial processes.<br>- Combustion of coal and petroleum generates $SO_2$. | The reactant between sulfur dioxide with water and air forms sulfuric acid, which is the main component of acid rain. Acid rain can cause deforestation. |
| **$O_3$** | - Reacting oxides of nitrogen with organic compounds in the presence of sunlight | Damaging the leaves of trees and other plants in addition to make the human breathing deeply and vigorously more difficult. |
| **CFC** | - Air conditioners.<br>- Refrigerators. | CFC can affect the lungs, central nervous system, heart, liver and kidneys. |

## 1.3 Objectives

The overall objective of the project is to develop a fixed wing UAV equipped with air quality sensors that collect data, transmit those collected data and provide a platform to see and visualize all the collected measurements in an easy and user-friendly fashion. In addition, it provides a live video streaming and supports autopilot feature through ground control station. This project is multidisciplinary between an electrical engineering team and a computer engineering team. Separation of concerns and work load division and synchronization were a major concern during the planning stage. Below is a description of each team's tasks and objectives.

Electrical Engineering Team:

a) Identify the UAV specifications.
b) Sensors selection optimization.
c) Implement air quality sensors and electronics for detecting real time data.
d) Make the power calculations and the decision on the power units needed.
e) Choose the suitable technology for transmitting data.

Computer Engineering Team:

a) Interfacing air quality sensors and transmitting their measurements.
b) Providing a live video stream.
c) Provide tools to compare, export and publish obtained results.
d) Provide a platform to store, view and visualize air quality data through website and android application.

## 1.4 Project Plan

In this section, the whole project main tasks including SDP1 and SDP2 are listed below and Gantt charts in Figure 1-2 and Figure 1-3 used to elucidate the time slot for the tasks.

The main tasks for Senior Design Project 1 are:

1- Setting the project constrains and standards.
2- Looking for previous related work “Literature review”.
3- Searching for different types of air quality sensors (CO, $CO_2$, NO, dust and humidity) in term of output type analog or digital, price, weight, number required of pins, etc.
4- Checking the efficiency for the solar cell that is available in campus and search for a suitable one if the available one does not give acceptable results.
5- Measuring the power consumption and indicate the effect of airflow on sensors.
6- Comparing several types of UAV planes and select the suitable one.
7- Checking if the sensors' operations are affected by high speed.
8- Measuring the power consumption of Arduino and Raspberry pi with extreme load.

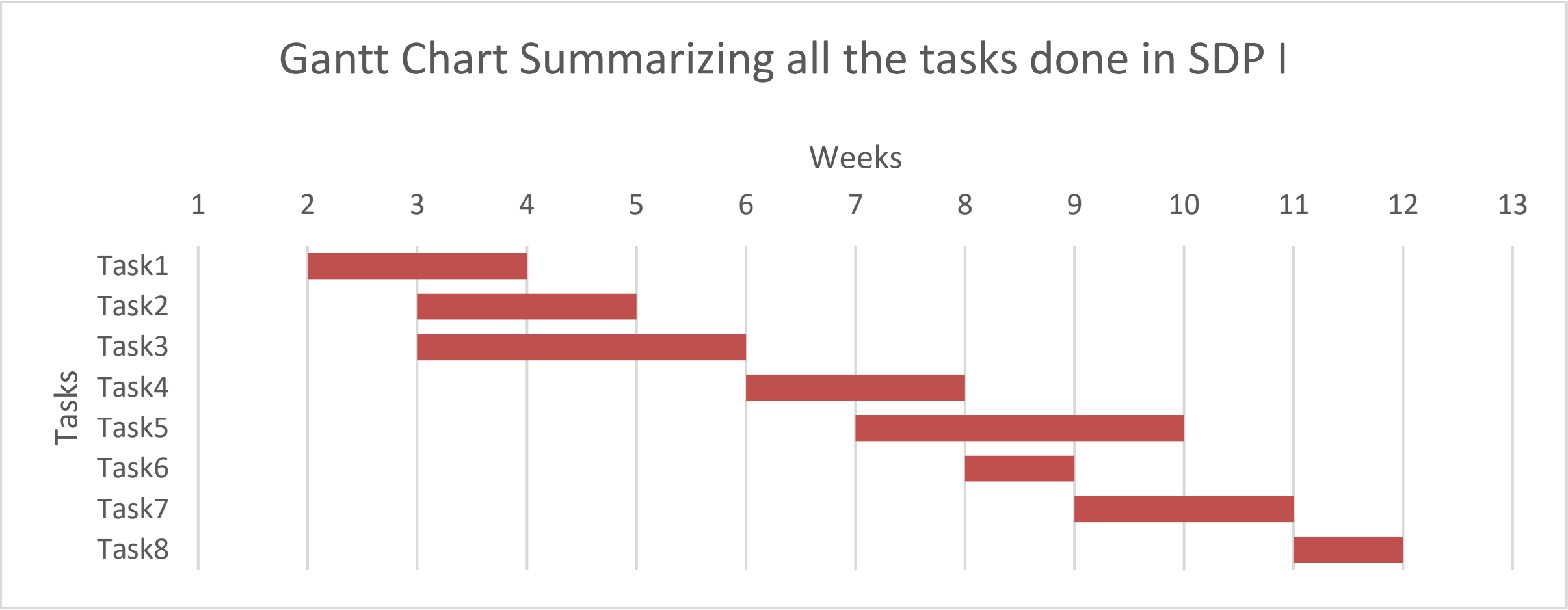

**Figure 1-1: Gantt chart for given tasks in SDP I**

The main tasks for Senior Design Project 2 are:

1- Checking the funding of the project.
2- Preparing a purchase list.
3- Doing practical implementing and testing for the sensors.
4- Experimenting the communication between the sensors and camera with the server.
5- Looking for a suitable Li-Po battery.
6- Flying the plane to test the payload.
7- Testing the sensors and communication during flight.
8- Modifying the frame of the plane to have a suitable place for sensors and other components.
9- Setting the autopilot parameters on the flight controller.
10- Flying the plane in auto-pilot mode with live streaming and all sensors working.
11- Doing report and presentation.

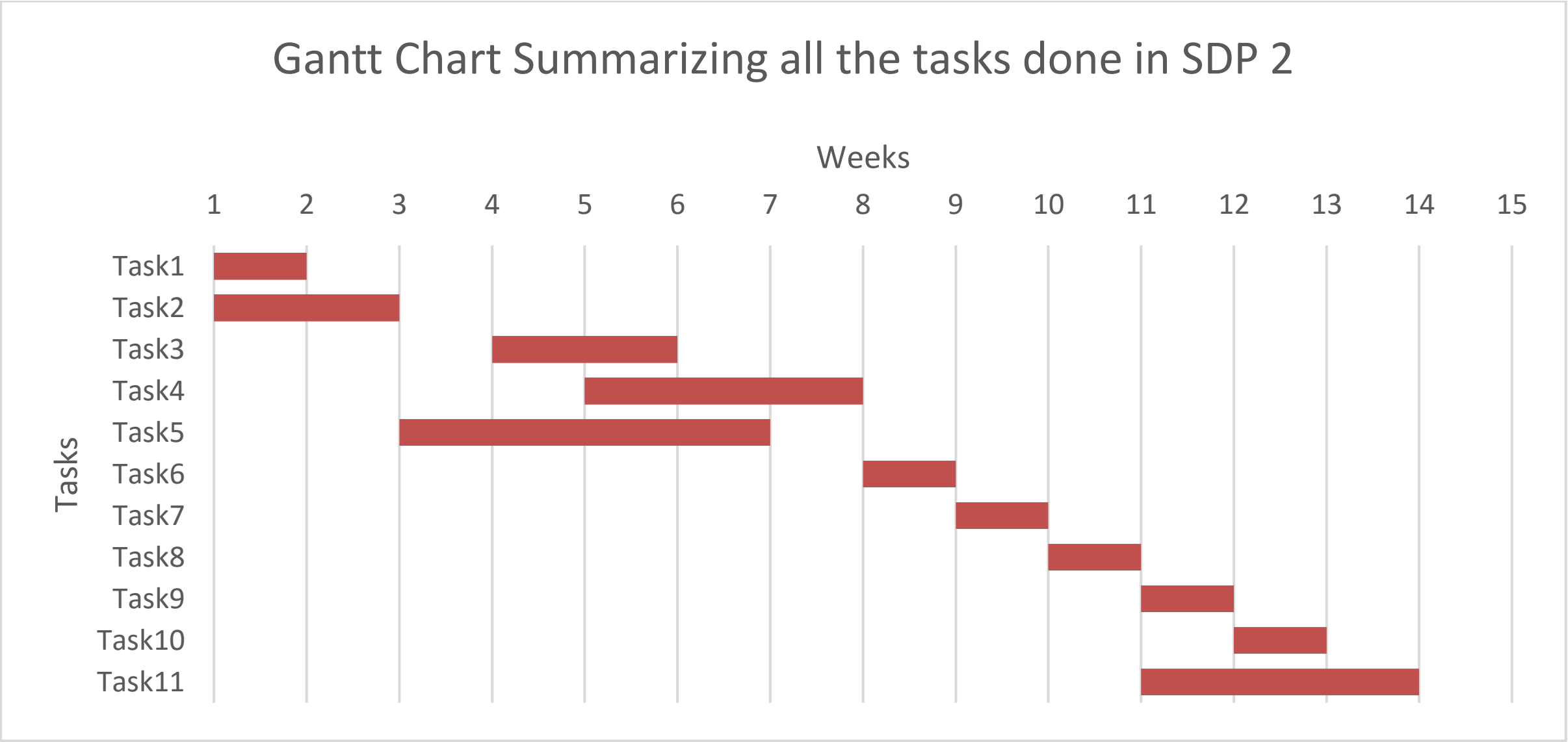

**Figure 1-2: Gantt chart for given tasks in SDP 2**

# Chapter 2 LITERATURE REVIEW

## 2.1 Related Works

We will now summarize the work that has been done in literature, providing a benchmark for us to realize our contribution and how our work compares and provides advantages to others.

### 2.1.1 Design an integrated solar powered UAV for remote gas sensing

This research focused on the use of UAVs as a tool to gather air quality parameters. As a proof of concept that has been used, the plane was a fixed wing UAV remotely controlled via a radio controller. A non-dispersive infrared sensor used to measure the gases and transmit them through a node-network back to the base. Using this node network, it was able to increase the range of the UAV. The UAV used in this paper was a Green Falcon Kit that has a wingspan of 2.52 meters and a payload of 3 kg. A flight time of 1 hour achieved by mounting a high capacity battery and integrating solar cells for increased power efficiency. The used sensors had an accuracy of ± 0.26ppm. An Ardupilot Mega 2.5 autopilot software installed on an APM flight controller. The Ardupilot Mega supports flight paths set prior to take off and then it can handle the flight path tracking autonomously as well as a manual override for emergencies where manual control can be forced. In case of loss of communication with the ground station, the flight controller software will autonomously return the UAV to point of takeoff and wait for further commands [3].

Similarly, in this project, a fixed wing UAV was being utilized with electrochemical sensors for air quality measurement and solar cells in order to produce more energy. In addition, remotely controlled setup was used while the paper above uses an autonomous path following autopilot. In this project, the system utilizes existing mobile phone network. On the other hand, the previous work utilizes a base of nodes on the ground to transmit data.

### 2.1.2 UAV application in ecology: Data collecting with quad-copter

This project utilized a user measurement system for monitoring air quality and air pollution. Quadcopter is the main core of the system and it is equipped with sensors to measure the following gases vertically and horizontally: $O_2$, $CO_2$, pressure, temperature, and humidity. The total weight is 1.4 kg with initial payload of the quadcopter. Using Arduino as a data acquisition device they pre-set the flight plan to navigate to specific coordinates via onboard GPS system and record measurements in local memory to be retrieved upon landing. The ground station which is operated by a human controller, and the quadcopter communicate wirelessly via 433 MHz link. Their setup was driven by a 5Ah 12V battery. The flight time of 20 mins and maximum velocity of 8 m/s. Control was made autonomous with a backup RC controller in case of GPS signal loss. The setup was tested and a data for $CO_2$ and $O_2$ in collected in different locations with different value of air pressure to find out the performance of the sensors during the air pressure. This system is considered to be fast, accurate and affordable measurement system to monitor and measure the level of $CO_2$ and $O_2$ that allowing ecologists to gather data fast and to analyze measured information [4].

### 2.1.3 SCENTROID DR300

SCENTROID DR300 is a commercial quadcopter sold along a variety of sensors (30 to choose from) all measuring air quality. Only two sensors can be equipped at any given time, the collected data along with live video are transmitted wirelessly via a 2.4 GHz link to an android application. Flight is controlled manually. Flight time is approximately 25 minutes. This product is commercially available and does not require much time to setup. It is built on the Phantom quadcopter sold by DJI Company. However, because only two sensors are supported at any given time and the fact that flight time is limited to 20 mins makes collecting larger amounts of data for all the air quality parameters more difficult [5].

### 2.1.4 Ambient air quality in Qatar: a case study on Ras Laffan Industrial city

The author conducted analysis of ground station data monitoring NOx and SO2 as well as CO2. This study aimed at providing a model to predict future values of pollutants. The followed approach in this study is different from the one to be followed in this report. However, the previous data will be used as reference values after implementing sensors. One interesting aspect in the study is the generation of heat maps that illustrated regions and area with higher pollution rates [6]. Computer engineering students inspired to follow the method in this study and embed an option in the website to automatically generate heat maps based on collected data.

### 2.1.5 Cellular for the skies: Exploiting mobile network infrastructure for low altitude air-to-ground communications

This paper focused on the use of 4G network [7] [8] for control and transmission of data from a UAV. In Qatar 3G coverage is available everywhere, 4G is available in most places and 4G+ is available in and around major cities. The delay time varies between 10-50ms theoretically, while experimentally the worst case was around 1.7 seconds. While this delay can cause serious problems with manual control of the flight [7]. However, this project will use the data of the mobile network to transmit the collected data that can handle the range of delay.

### 2.1.6 Related work comparison

The table below shows the related work compared with this project.

**Table 2-1: Related work comparison table**

| Project No. | Sensors | Type | Flight Time | Control | Real time transmission |
|---|---|---|---|---|---|
| **2.1.1 [3]** | $CH_4, CO_2$ | Fixed wing | 1 hour | Pre-set coordinates | No |
| **2.1.2 [4]** | $O_2, CO_2$ | Quadcopter | 20 mins | Pre-set coordinates, RC control | No |
| **2.1.3 [5]** | Many | Quadcopter | 25 mins | RC control | Yes |
| **2.1.4 [6]** | $No_x, CO_2, S$( | Ground Station | N/A | N/A | No |
| **This project** | Many | Fixed wing | 20 mins | RC/ Pre-set coordinates | Yes |

## 2.2 UAV Types

UAVs are classified into two types: fixed wing and rotary wing. Fixed wing airplanes are comparative in configuration to planes used in human and cargo transportation. It is the constant forward development produced by the turn of a propeller which lifts these units off the ground and gives their capacity to maintain flying. The basic outline finish with one unbending wing across the body takes into consideration high speeds and long distance travelling. This sort of automation is perfect for scope of huge territories, for example, surveillance and mapping applications. Some fixed wing units additionally make the ideal eye-in-the-sky for farmers. On the other hand rotary wing UAVs are much similar to manned helicopters, Similarly they have multiple propellers that lift the UAV and guide it in the desired direction, they can have from a single blade to 6 or even 8, each configuration is different however the principle is the same. Rotary wing drones are excellent for tasks that require the UAV to stay still in one place or move in a limited area. However, they can have a limited range of travel [8].

The table below gives a briefing about the differences between Fixed Wing UAVs and Rotary Wing UAVs (Quadcopter) in term of advantages and disadvantages [8].

**Table 2-2: Comparison between fixed wing and rotary wing UAVs**

| | **Fixed Wing** | **Rotary Wing** |
|---|---|---|
| **Advantages** | - High speed<br>- Long distance traveling<br>- Heavy payload<br>- Smooth gliding through air (sleek structure)<br>- Minimal maintenance process | - Multi directional (can hover)<br>- No need for runway to takeoff or land |
| **Disadvantages** | - One directional (cannot hover)<br>- Runway or launcher is needed to takeoff and landing<br>- Commercial use is restricted by rules and regulations | - Complex design<br>- Low operational time<br>- High cost<br>- Very light payload |

In this project, a fixed wing UAV has been chosen because it provides simpler aerodynamics and controls which increases the speed of the UAV. This speed increase can facilitate longer flight time and make it cover more distance thus we will get reading from more areas. Moreover, a fixed wing planes are more capable to carry more weight which is significant in this project.

## 2.3 Design Constraints

Engineering designing is all about creating solutions for real time problems. Through the design steps some limitations could be faced in order to achieve the objectives. To compromise between the design objective and the limitation some constraints have to be specified in advance. We construct a platform to justify our choices for each component in the system. Our objective in this project is to design and implement a UAV for air quality monitoring, the most general constraints we consider in this design are safety, weight, and cost.

The table below illustrating the design constraints we identified with a description of their application in our project.

**Table 2-3: Design constraints**

| Name | Description |
|---|---|
| **Communication** | a. Coverage<br>The coverage should be include all areas in Qatar so that the UAV can travel long distance without losing contact with ground station, At minimum of 20 km from the ground station. |
| | b. Data rate<br>Our system will support live video streaming over 4G, therefore high data rates to support quality stream should be available. The minimum data rate of 0.5 Mbps is required for a standard definition stream. |
| **Sensors** | a. Response time<br>Our UAV will be traveling at a higher speed of approximately 100km/h; therefore, the response time for sensor should be less than 10 seconds to provide high spatial and sampling frequency which adds accuracy in our system. |
| | b. Heating time<br>Some sensors available with vendors require a long period for heating up before they can start measuring data, for ease of use, the maximum heat up time is 10 minutes. |
| | c. Accuracy<br>The selected sensors should provide high accuracy so that the collected data would be useful for any environmental researchers, we recommend a +- 5% threshold. |
| **Plane** | a. Wings Area<br>Should be able to support the lifting of the required weight of the component. Wingspan should not be less than 1.5 m. |
| | b. Battery<br>The rechargeable battery should support UAV for an average flight of 20 min so that a single flight can cover approximately 20~30 km with maximum weight of 1.5 kg. |
| | c. Payload:<br>The payload of the UAV should be able to handle the UAV's propulsion system, sensors, and electronic components as well as the battery, not less than 2kg. |
| **Safety** | a. Control<br>The flight controller system should be able to handle loss of communication and return to point of origin until it establishes communication again.<br>b. Energy source<br>The UAV should be running at electrical power, so that in case of a catastrophic failure the UAV will not contain a flammable liquid that could cause major damage to humans.<br>c. Installation<br>The UAV should properly be constructed without loss components that can be effected by airspeed and fall over. |
| **Size & Weight** | The payload of the UAV should be able to handle the UAV's propulsion system, sensors, and electronic components as well as the battery. The maximum payload should not exceed 3 Kg. Therefore; The selected components should fit within the compartment of UAV and not greatly affect the overall weight of the UAV body so that the flying time will not decrease due to the increased payload. Payload |
| **Cost** | The project will be funded from Boeing company with a budget of $50,000. However, the overall cost should be in acceptable range. |

The table below shows some constraints and ways to match these constraints in the design process.

**Table 2-4: Constraints and solution**

| **Constraints** | **How to match the constraints** |
|---|---|
| **Communication**<br>**(Coverage, data rate, delay)** | 4G/ 3G dongle |
| **Sensors**<br>**(Accuracy, response time, heating time)** | Using of high quality sensors |
| **Payload** | Larger wing area |
| **Weight** | Choosing a lower weight materials and components |

## 2.4 Design Standards

The following table shows the design standards followed in this project.

**Table 2-5: Design Standards**

| Type | Name | Description |
|---|---|---|
| **WHO [9]** | World health organization | The measured results of contamination will follow these standards. |
| **LTE [10]** | Long Term Evolution | The data and the video is transmitted from the airplane to the server using LTE modem. |
| **USB 2.0 [11]** | Universal Serial Bus | The transmission of data between the Arduino to raspberry pi is done using USB. |
| **TCP/IP [12]** | Transmission Control Protocol/ Internet Protocol | The data transmission between the UAV and the ground station is done using TCP/IP protocol. |
| **HTTPS [13]** | Hyper Text Transfer Protocol Secured | The data is transmitted to the user over HTTPS protocol. |
| **GPS [14]** | Global Positioning System | The UAV is equipped with a GPS transmitter to identify its current location. |
| **H.264 [15]** | MPEG-4 Part 10 Advanced video coding | Video streaming in encoded using this standard. |
| **MAVLINK** | Micro Air Vehicle Protocol | A protocol used for communication between an unmanned vehicle and a ground station to exchange current locations and speed and receive commands, communication between the Pixhwak flight controller and ground station is done via this protocol. |
| **UDP** | User Datagram Protocol | The command forwarding between the Ground station and Flight controller via the Raspberry PI is done via this protocol. |
| **NMEA** | National Marine Electronics Association | The GPS receiver follows the NMEA standard to represent current location as well velocity and time. |

## 2.5 Design Assumptions

The following assumptions have been made:

1. We will be allowed to fly the UAV in predefined areas.
2. The computer engineering team will choose the appropriate method to interface the sensors and transmit data.
3. Network coverage (3G, 4G or 4G+) will be available throughout the flight path.
4. There will be a reasonable delay in the transmission of the video captured by the UAV.

# Chapter 3 SYSTEM OVERVIEW

In this chapter, the system Block diagram of the overall system will be described with a principle of working.

## 3.1 Block diagram

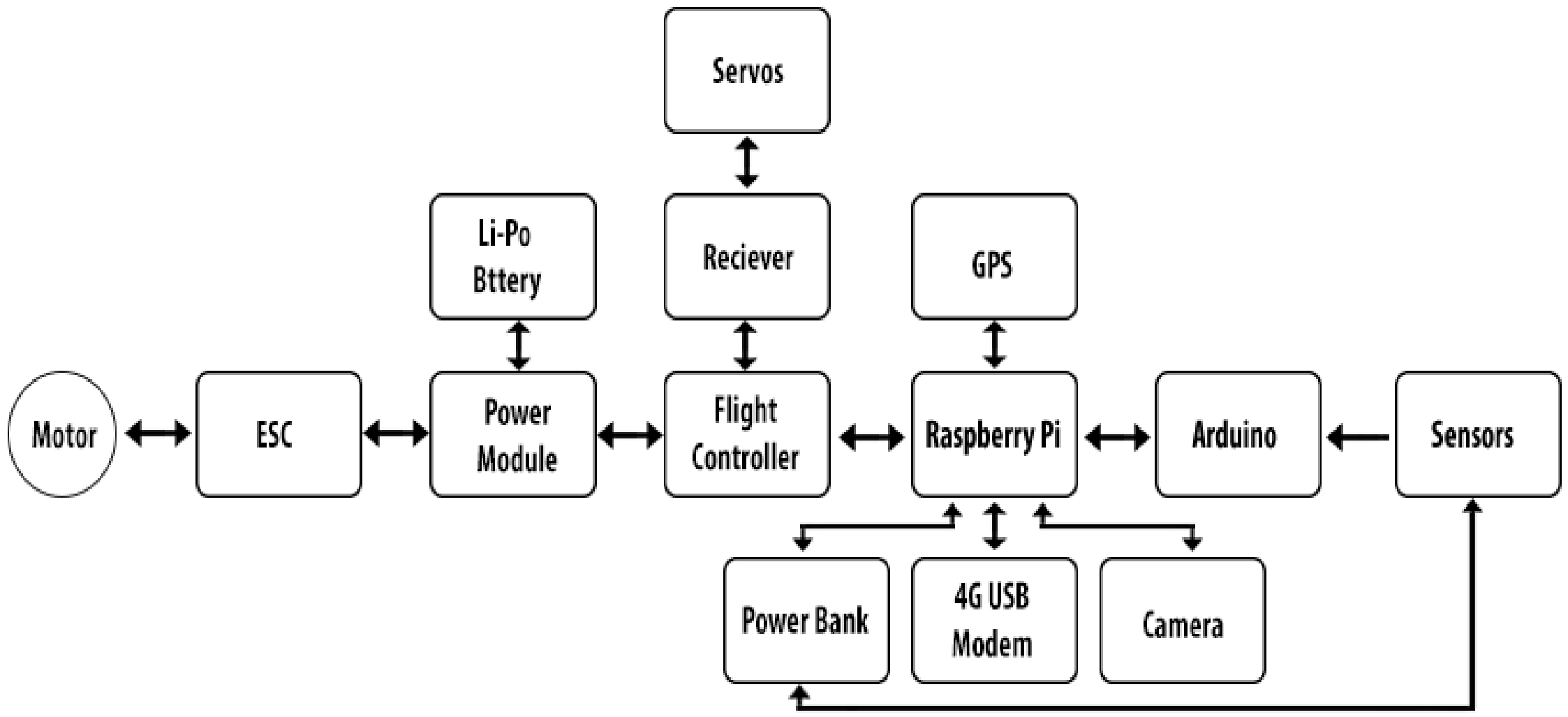

**Figure 3-1: System block diagram**

Figure 3-1 shows that the UAV module will consist of sensors with interfacing circuitry, Arduino for data collection, Raspberry Pi for data transmission via 4G modem, GPS to link measurement to a location and height where it taken at, Camera for live streaming and an open source flight controller to support installation of autopilot software

### 3.1.1 Motor

The UAV will include a brushless DC motor that runs on DC electric power supplied by a battery in order to convert the electrical energy into mechanical torque. The advantages of brushless DC motors are numerous, Very precise speed control, high efficiency, high reliability, reduced noise, longer lifetime (no brush abrasion) and No ionizing sparks.

### 3.1.2 Battery

The UAV will operate with a 5,300 mAh 6S Lithium Polymer (Li-Po) battery. Lithium Polymer batteries comes with a wide array of advantages such as lighter weight, higher capacity and higher discharge rates compared with other batteries. In the battery, the main two parameters are the battery voltage and number of cells. The voltage provided by the battery must be at least equal to the voltage of the motor in order to operate. On the other hand, the capacity of a battery, which is a measurement of how much power the battery can hold. In other word, it represents how much current a battery will discharge over a period of one hour knowing that more capacity will lead to higher weight of the battery.

### 3.1.3 Electronic Speed Control

An electronic speed control (ESC) is an electronic circuit that control the speed of an electric motor and its direction. An ESC has two main sets of wires.  One lead will plug into the main battery. The second lead power the brushless motor.

### 3.1.4 Flight Controller

The Pixhawk flight controller is an open hardware project that bundles a processer with the necessary sensors to operate a flight, the package includes 168 MHz CPU with 256 KB Ram as well as a Gyro, compass and 3D accelerometer and all the different connectors to connect to the plane. This open platform supports the installation of autopilot software which was used, Autopilot. Along with its stability and reliability, it fulfills the needed requirement.

### 3.1.5 Sensors

The UAV will carry sensors that will monitor the air quality in Qatar. Theses sensors (CO, CO2, O3, dust, temperature and humidity) will be powered from the power bank and they will be interfaced with a microcontroller to transmit data.

### 3.1.6 Arduino

Arduino is an open-source electronics platform that makes hardware and software easy to use. The board is able to read inputs from a sensor and turn it to output. Nowadays, Arduino is available with different processors, sizes and price and easy to get them.

In this project, the main functionality of Arduino Nano is to interface with sensors, and read data from the connected sensors and send them to Raspberry Pi through USB serial port.

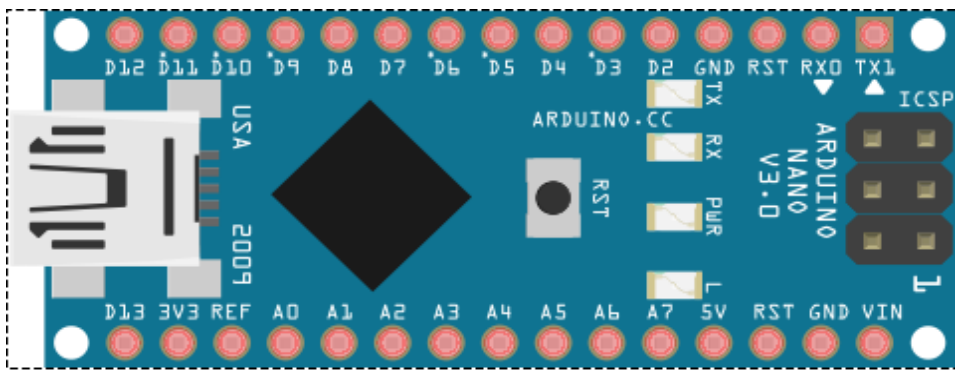

**Figure 3-2: Diagram of a typical Arduino Nano board [16]**

### 3.1.7 Raspberry pi

Raspberry pi is a small computer that can be used in electronic project it is similar to PC. In this project will use Raspberry pi for connecting camera for online streaming, GPS for location tracking and to communicate between UAV and server through 4G/3G USB. Raspberry pi came with several models, the pi 1 A+ model the lower cost one with 512MB RAM, 1 USB port and no Ethernet port. Pi 1 B+ model it has 512RAM, 4 USB port and Ethernet port. The Raspberry Pi 2 Model B is the second generation and it replaced the Raspberry Pi 1 Model B+. The features of Raspberry Pi 2 Model B are 900MHz quad-core and 1GB RAM [17].

Raspberry pi 3 is the third-generation Raspberry Pi. It has 1.2GHz 64-bit quad-core and It will be used in this project based on computer engineering student recommendation because it is capable of doing multiple tasks at a time. In addition, Raspberry Pi is much faster and has larger memory than Arduino. It has video output for streaming and Internet can easily be run on Raspberry Pi using 4G USB dongle.

### *3.1.8* Camera

The camera will be used when there is a sudden increase in a specific gas or in emergency cases. For example, if certain limit of gas increased or a fire is happened, the webcam will record the source of this place.

### 3.1.9 3G/4G USB Modem

One of the major part in this project is how to send real time data of the sensors from the plane to the ground. The 3G/4G dongle will play the role of transmitting data that gathered from sensors and webcam to the ground.

## *3.2 Data viewing module*

The Data viewing module provides two tools to view and visualize the collected data for the convenience of the system users; these tools are a website and android application that can be used to show data in any of the following formats:

- Traditional tabular form.
- Graph (for any time span chosen).
- Google map showing collected measurements.

### 3.2.1 Website

Within the website there are the viewing options mentioned above. Upon selecting a viewing option, the user will choose which parameters he wishes to see and generate any graph (colored google map, or traditional graph) as well as video live stream of the UAV.

### 3.2.2 Android Application

The Android application has similar functionalities to the website, they can be used interchangeably to achieve the same goal. Upon login, the user can select which parameters he wishes to see and graph (colored google map, or closest geographically recorded values) as well as video live stream from the UAV.

# Chapter 4 SELECTION OF SYSTEM COMPONENTS

## 4.1 *UAV module*

**Table 4-1: UAV module comparison**

| **Plane Name** | Fun World 3D | ZEN-120 | Stick-60 |
|---|---|---|---|
| **Code No.** | A134R-PUR | A112-BLU | GPMA1221 |
| **Wing Span** | 1790 mm | 1800mm | 1740 mm |
| **Wing Area** | 8980 $cm^2$ | 6160$cm^2$ | 6400 $cm^2$ |
| **Flying Weight** | 4300g | 4500g | 2900-3400 g |
| **Fuselage Length** | 1780mm | 1820mm | 1420 mm |
| **Engine Required** | Rimfire 1.20 Brushless electric | Rimfire 1.20 Brushless electric | XM6350EA-6 |
| **Radio Required** | 8 channels, 8 servos | 5 channels, 5 servos | 4 channels, 4 servos |
| **Approx. cost** | $1420 | $1400 | $1300 |

The Stick-60 has been chosen based on different kind of criteria such as the wingspan, wing area, aspect ratio, motor, propeller and cost. However, the power consumption play an important role in the selection since it will affect the flying time.

### 4.1.1 Selected UAV parameter calculation

There are many constraints in the operation of choosing the airplane, and based on some analysis, the best airplane will be selected. Normally, the plane has 4 acting forces on it which will be described in the Figure 4-1 below [18]

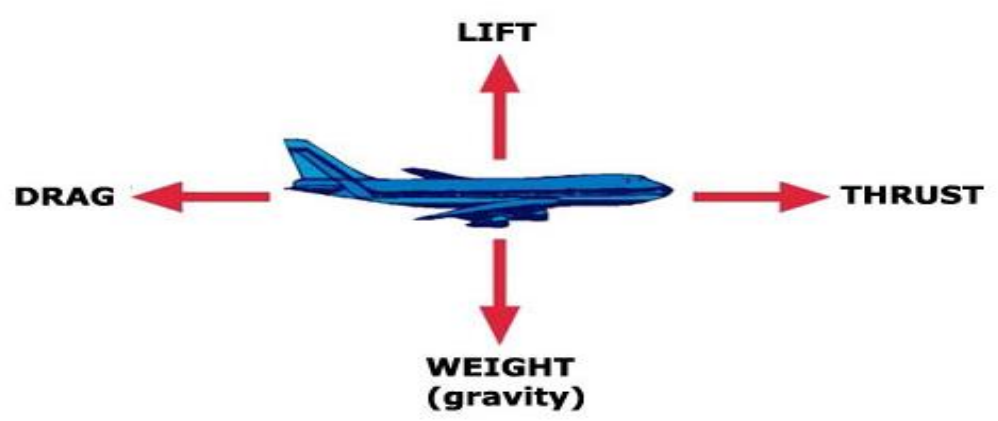

**Figure 4-1: Forces acting on the airplane**

In order to allow the plane to raise up the, lift force must be greater than the weight force, also for the plane to move forward, the thrust force must be greater than the drag force [18]. To meet this limitation, the process needs some engineering analysis. In order to generate lift more than weight, either reduce the weight or increase the lift force. However, the frame weight of the UAV could not be reduced due to the manufacturing of the airplane and readymade piece. Therefore, the only solution is to increase the lifting force. Referring to the mechanical analysis, the lift force affected by some parameters such as density of air around the wing, wing area, the needed velocity of UAV, airfoil coefficient and angle of attack. For the plane to accelerate vertically, the ratio between thrust and the weight must be greater than one [18], so in the simplest way:

$$\frac{Thrust}{Weight} = \frac{a}{g} = \frac{Ft}{Fw} \geq 1 \tag{1}$$

High F/w indicates high vertical acceleration. The ascent stage will be ignored due to the high complexity of the calculation. That might affect the exact calculation of the total power consumption.

The equation below shows the minimum velocity when the left force equal weight force.

$$Fw = Fl = (0.5\times C_l \times \rho air \times V^2 \times A) \tag{2}$$

This equation is relating the weight with the lift parameter (lift coefficient, air density, velocity and surface area of the wing).

Based on the airfoil design in the airplane datasheet, the wing shape is semi-symmetrical [19]. Therefore, it follows NASA standers from NACA2408 to NACA2415 [20]. $C_L$ is a variable coefficient has a typical range between 0.6 and 1.2, and it will equal to one based on the estimation that proposed.

To get the value of the velocity, assuming the lift force equal to the weighted force

$$Fw = Fl = gravity \times mass \tag{3}$$

$$V = \sqrt{\frac{2\times Fw}{C_l \times A \times \rho\text{air}}} \tag{4}$$

Where weight force of the plane is the total weight of the plane which is 4kg multiplied by gravity (~9.81 $m/s^2$) which equals approximately to 40N. The air density is approximately equal to 1.225 kg/m3 [21], and the wing area is $0.64m^2$

$$V = \sqrt{\frac{2\times 40}{1\times 0.64\times 1.225}} \cong 10.1\frac{m}{s}$$

The airplane is case sensitive to the payload and from the equation the $V \propto mass$. Depending on the equation, bigger wing surface area allow the UAV to take off at low velocity compared with a smaller wing surface area that needs higher velocity to take off as shown in Figure 4-2.

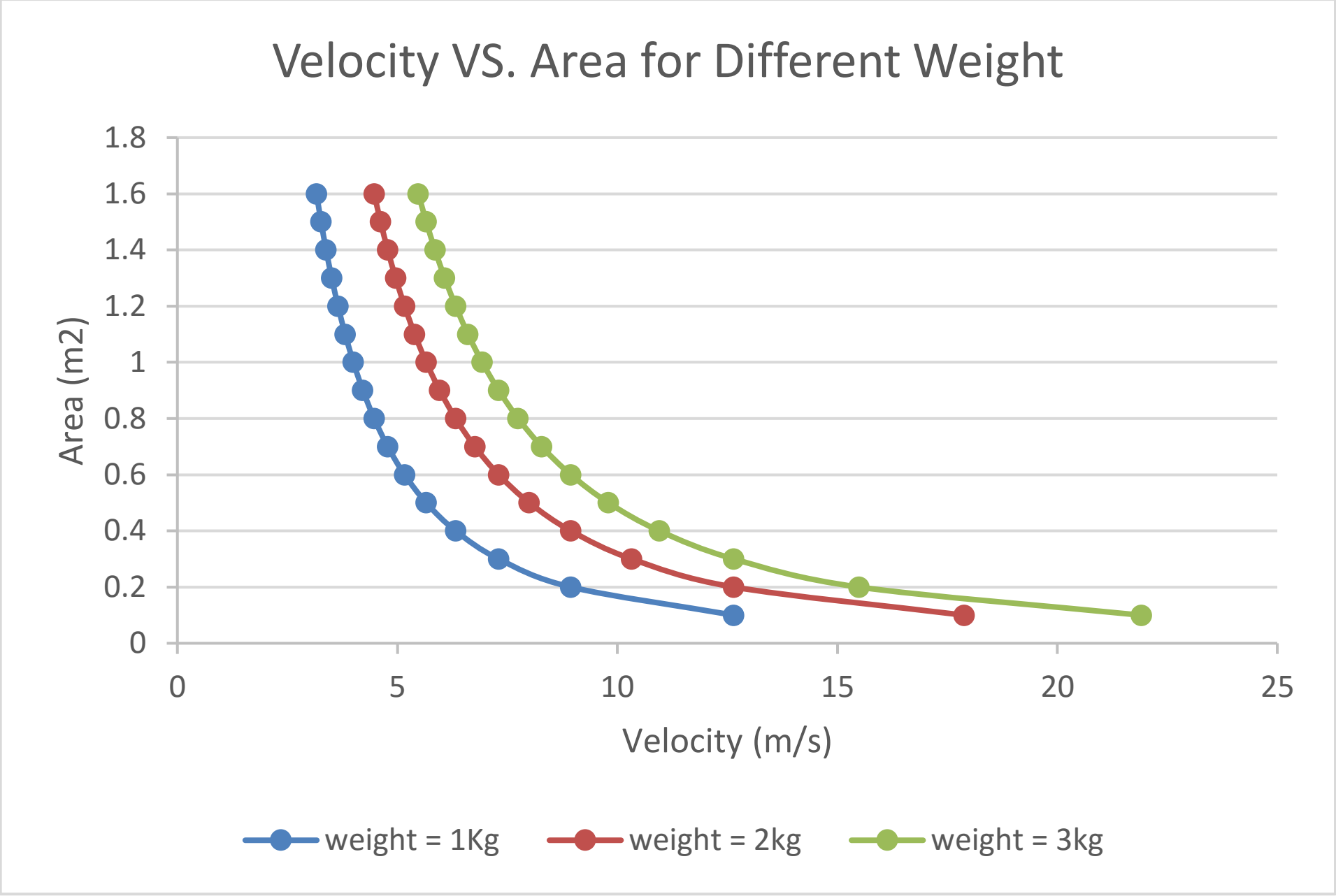

**Figure 4-2: Take-off velocity Vs Area for different weight**

The real lift coefficient is determined based on the Reynold number in relation to the dimension of the wing.

$$Re = \frac{V \times W}{\gamma} = \frac{10.1 \times 0.3}{1.5111 \times 10^{-5}} = 200{,}529 \quad (5)$$

V: is the ground speed in m/s.

W: wing width (chord in this case) in meters.

ɣ: is the kinematic viscosity (m2/s)

Referring to NACA standards, this airfoil indicate the change in lift coefficient $(C_l)$, drag coefficient $(C_d)$ and angle of attack (α) based on the Reynold number.

The exact equivalent coefficient of lift $(C_l)$ and coefficient of drag $(C_d)$ and angle of attack (α) are founded to be 1.02, 0.02, 11° respectively [22].

In other words, if the initial velocity known, the initial lift coefficient could be obtained also from the above equation by fixing the value of the velocity to 10.1m/s.

$$C_l = \frac{F_w}{0.5 \times \rho air \times V^2 \times A} \quad (6)$$

$$C_l = \frac{40}{0.5 \times 1.225 \times 10.1^2 \times 0.63} \cong 1$$

From the above equation, the velocity of the plane is ∝ to the $C_l$ Thrust force is the force that moves the airplane forward. So, In order to calculate this force, some parameters are needed to be included and take care of it such as propeller size and motor speed. The thrust force could be simplified using this equation [23].

$$F_t = 4.3294399\times10^{-8}\times RPM\times\frac{d^{3.5}}{\sqrt{pitch}}(4.3294399\times10^{-4}\times RPM\times pitch - vo) \quad (7)$$

Where Ft is dynamic thrust (it is called static thrust if V0 = 0), measured in N, RPM represent the airplane rotation per minute, pitch is propeller pitch in (inches), d represent the propeller diameter in (inches) and $V_o$ is the forward velocity measured, in (m/s).

The RPM of the recommended motor is obtained from the below:

$$RPM = Kv\times volt \quad (8)$$

$$\mathrm{RPM} = 22.2\times560 = 12432$$

Then substitute in equation (7),

$$F_t = 4.3294399\times10^{-8}\times12432\times\frac{14^{3.5}}{\sqrt{6}}(4.3294399\times10^{-4}\times12432\times6 - 10.1) \cong 50N$$

By changing the values of the velocity and fixing the propeller input, the thrust Vs airplane speed will be as shown below

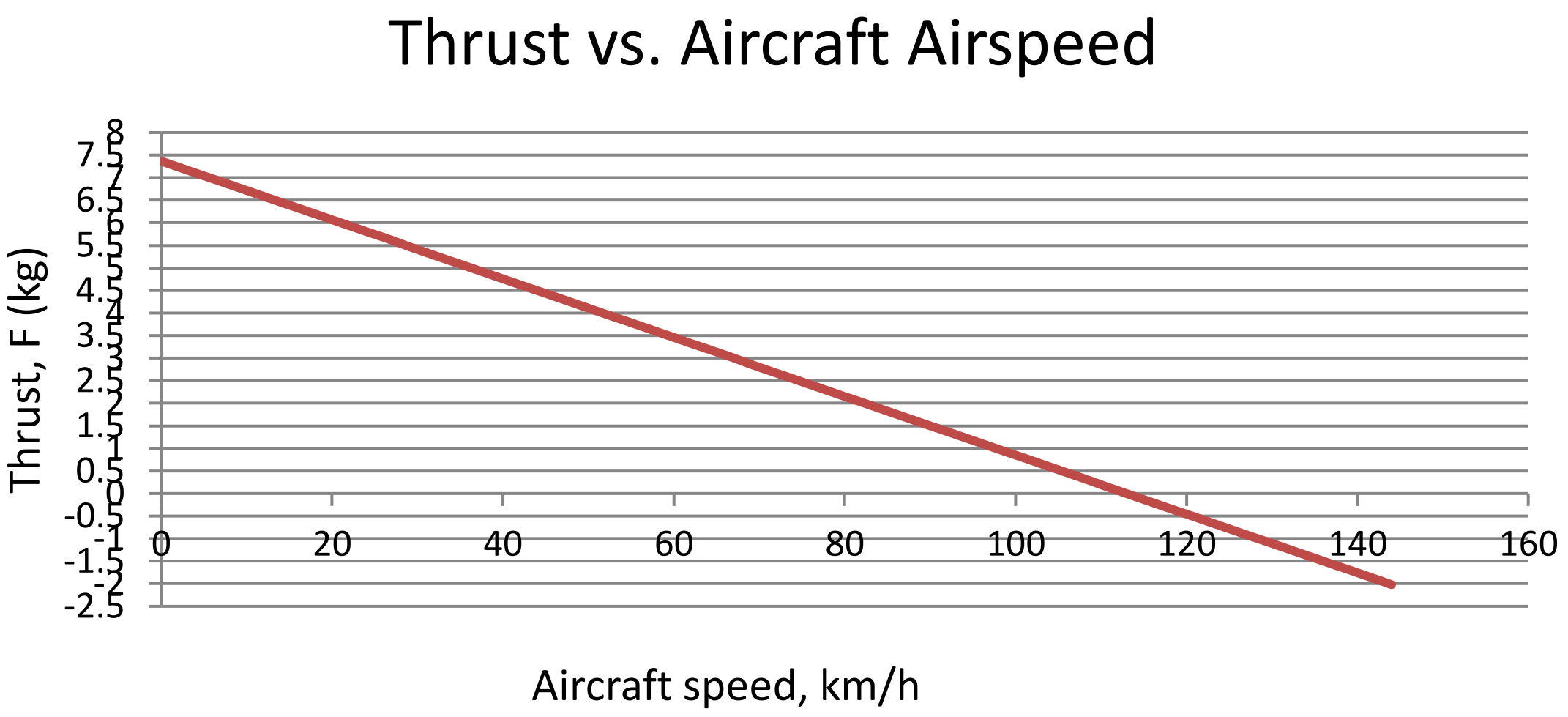

**Figure 4-3: Thrust Force (kg) vs speed in ( Km/h)**

From Figure 4-3, if the weight is 4 kg, the velocity of the UAV will be around 60 km/h.

In order to maintain the thrust force is greater than the weighted force, the maximum weight should not exceed 5 kg.

To calculate the net force along x-axis ($F_x$)

$$\sum F_X = F_T - F_d - F_r \quad (9)$$

Where $F_d$ is the drag force and $F_r$ is the rolling friction force.

$$F_d = 0.5\times C_d\times \rho air \times V^2 \times A \quad (10)$$

$$F_d = 0.5\times0.02 \times 1.225 \times 0.63\times 10.26^2 = 0.81\ \mathrm{N}$$

$$F_r = \mu \times \mathrm{Fw} \quad (11)$$

$$\mathrm{F_r} = 0.005 \times 40 = 0.2\ \mathrm{N}$$

Where µ is the coefficient of sliding friction, (assume to be small in flat road airplane approximate to 0.005 [24].

$$\sum F_X = F_t - F_d - F_r \quad (12)$$

$$\sum F_X = 50 - 0.81 - 0.2 \cong 49\ N$$

Getting the acceleration calculation

$$\sum F_X = m \times a \quad (13)$$

$$a = \frac{Fx}{m} = \frac{49}{4} = 12.25\,\frac{m}{s^2}$$

Now, the mechanical power of the motor could be calculated using this equation below:

$$P_{mech} = F_X \times V \quad (14)$$

Based on the force resulted from the thrust force and the velocity, the mechanical power could be calculated from equation 14

$$P_{mech} = 49 \times 12.25 \cong 600\ Watt$$

The efficiency of the motor is 92% so the electrical power

$$P_{elec} = \frac{\mathrm{P_{mech}}}{\eta} \quad (15)$$

$$P_{elec} = \frac{600}{0.92} = 652\ watt$$

Notice that the power has a relation with the force and the force is related to many parameters especially the weight. In order to optimize the power consumption, the propeller design and larger wing surface area are key factors in decreasing the velocity and sequentially decrease the power consumption.

### 4.1.2 Web site real values simulation

In this part, further checking process was implemented using real values simulator via online website [25]. Some parameters are needed such as the wing area, battery specification, motor type, speed controller, propeller, air temperature, pressure in order to get all the required data regarding the flight time, maximum weight, currents, voltages and static thrust.

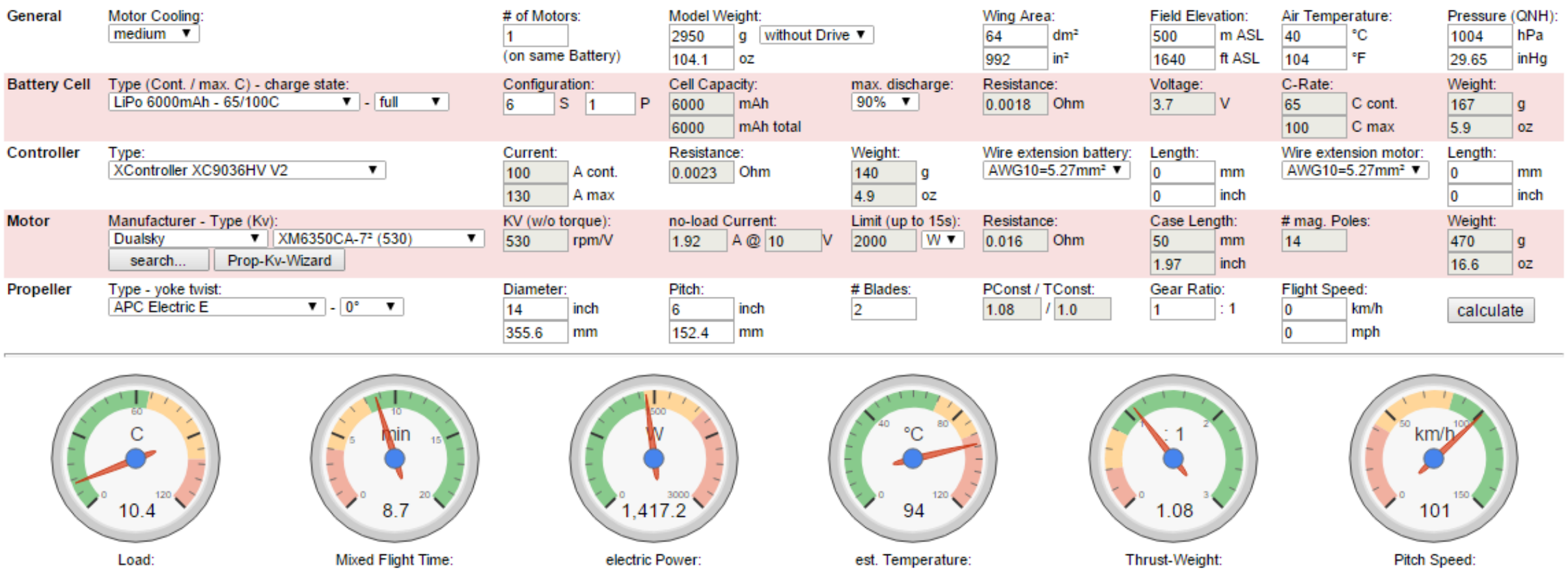

**Figure 4-4: Input parameters**

| Battery | |
|---|---|
| Load: | 10.40 C |
| Voltage: | 22.86 V |
| Rated Voltage: | 22.20 V |
| Energy: | 133.2 Wh |
| Total Capacity: | 6000 mAh |
| Used Capacity: | 5400 mAh |
| min. Flight Time: | 5.2 min |
| Mixed Flight Time: | 8.7 min |
| Weight: | 1002 g |
| | 35.3 oz |

| Motor @ Optimum Efficiency | |
|---|---|
| Current: | 62.54 A |
| Voltage: | 22.71 V |
| Revolutions*: | 11052 rpm |
| electric Power: | 1420.5 W |
| mech. Power: | 1274.8 W |
| Efficiency: | 89.7 % |

| Motor @ Maximum | |
|---|---|
| Current: | 62.39 A |
| Voltage: | 22.71 V |
| Revolutions*: | 11054 rpm |
| electric Power: | 1417.2 W |
| mech. Power: | 1271.8 W |
| Efficiency: | 89.7 % |
| est. Temperature: | 94 °C |
| | 201 °F |

| Wattmeter readings | |
|---|---|
| Current: | 62.39 A |
| Voltage: | 22.86 V |
| Power: | 1426.2 W |

| Propeller | |
|---|---|
| Static Thrust: | 5108 g |
| | 180.2 oz |
| Revolutions*: | 11054 rpm |
| Stall Thrust: | 4617 g |
| | 162.9 oz |
| exc.Thrust @ 0 km/h: | 5108 g |
| exc.Thrust @ 0 mph: | 180.2 oz |
| Pitch Speed: | 101 km/h |
| | 63 mph |
| Tip Speed: | 741 km/h |
| | 460 mph |
| specific Thrust: | 3.60 g/W |
| | 0.13 oz/W |

| Total Drive | |
|---|---|
| Drive Weight: | 1773 g |
| | 62.5 oz |
| Power-Weight: | 311 W/kg |
| | 141 W/lb |
| Thrust-Weight: | 1.08 : 1 |
| P(in) @ max: | 1468.2 W |
| P(out) @ max: | 1271.8 W |
| Efficiency @ max: | 86.6 % |

| Airplane | |
|---|---|
| All-up Weight: | 4723 g |
| | 166.6 oz |
| Wing Load: | 74 g/dm² |
| | 24.2 oz/ft² |
| Cubic Wing Load: | 9.2 |
| est. Stall Speed: | 42 km/h |
| | 26 mph |
| est. Speed (level): | 93 km/h |
| | 58 mph |
| est. Speed (vertical): | 8 km/h |
| | 5 mph |
| est. rate of climb: | 8.8 m/s |
| | 1738 ft/min |

**Figure 4-5: Real output values**

From Figure 4-5, the maximum flight time founded to be around 10 min in the maximum load, the electrical power is 1000 watts, the thrust force is 4650g which near to the calculated results above.

| Propeller | Throttle | Current (DC) | Volage (DC) | el. Power | Efficiency | Thrust | | Spec. Thrust | | Pitch Speed | | Speed (level) | | Motor Run Time |
|---|---|---|---|---|---|---|---|---|---|---|---|---|---|---|
| rpm | % | A | V | W | % | g | oz | g/W | oz/W | km/h | mph | km/h | mph | (90%) min |
| 1600 | 13 | 0.4 | 23.5 | 9.2 | 42.0 | 107 | 3.8 | 11.6 | 0.41 | 15 | 9 | - | - | 827.7 |
| 2400 | 20 | 0.9 | 23.5 | 21.3 | 61.1 | 241 | 8.5 | 11.3 | 0.40 | 22 | 14 | - | - | 356.7 |
| 3200 | 27 | 1.8 | 23.5 | 42.5 | 72.5 | 428 | 15.1 | 10.1 | 0.35 | 29 | 18 | - | - | 178.7 |
| 4000 | 34 | 3.2 | 23.5 | 76.0 | 79.3 | 669 | 23.6 | 8.8 | 0.31 | 37 | 23 | - | - | 99.9 |
| 4800 | 41 | 5.3 | 23.5 | 124.9 | 83.4 | 963 | 34.0 | 7.7 | 0.27 | 44 | 27 | - | - | 60.7 |
| 5600 | 48 | 8.2 | 23.4 | 192.4 | 85.9 | 1311 | 46.2 | 6.8 | 0.24 | 51 | 32 | - | - | 39.3 |
| 6400 | 55 | 12.1 | 23.4 | 282.0 | 87.5 | 1712 | 60.4 | 6.1 | 0.21 | 59 | 36 | - | - | 26.8 |
| 7200 | 63 | 17.1 | 23.3 | 396.8 | 88.6 | 2167 | 76.4 | 5.5 | 0.19 | 66 | 41 | 59 | 36 | 19.0 |
| 8000 | 70 | 23.3 | 23.3 | 540.2 | 89.2 | 2675 | 94.4 | 5.0 | 0.17 | 73 | 45 | 68 | 42 | 13.9 |
| 8800 | 78 | 31.0 | 23.2 | 715.7 | 89.7 | 3237 | 114.2 | 4.5 | 0.16 | 81 | 50 | 74 | 46 | 10.4 |
| 9600 | 85 | 40.3 | 23.1 | 926.8 | 89.9 | 3853 | 135.9 | 4.2 | 0.15 | 88 | 55 | 81 | 50 | 8.0 |
| 10400 | 93 | 51.5 | 23.0 | 1176.9 | 90.0 | 4521 | 159.5 | 3.8 | 0.14 | 95 | 59 | 88 | 55 | 6.3 |
| 11054 | 100 | 62.4 | 22.9 | 1417.2 | 89.7 | 5108 | 180.2 | 3.6 | 0.13 | 101 | 63 | 93 | 58 | 5.2 |

**Figure 4-6: Motor partial load**

From Figure 4-7, we can conclude that the flying time depends throttle percentage. For instance, if the motor fully loaded we notice that the rpm reach its maximum value and the current will reach 62 A. however this will reduce the motor run time, and this can be noticed especially during takeoff.

## *4.2 Sensors*

### 4.2.1 Sensor classification

For gas detection, there are a wide variety of devices based on different kind of materials and the principles of operations. Gas sensors are classified into many kind of sensors type; various approaches can be used. For example, (1) optical sensors, (2) electrochemical sensors, (3) electrical sensors, (4) mass-sensitive sensors, (5) calorimetric sensors, and (6) magnetic sensors. In this project, the sensors will be chosen based on the desired range and constrains of the design such as accuracy, heating time and response time. For instance, electrochemical sensor is being used in gases, dust is an optical sensor, temperature and humidity sensor is an electrical sensor.

### 4.2.2 Sensors measurement standards

In order to measure the air quality in Qatar, the results that will be collected from the sensors should be in a specific range. The detection range of the sensor must follow standards, and the limit of them should not be exceeding that.

The units of measure for the standards are parts per million (ppm) by volume, parts per billion (ppb) by volume, and micrograms per cubic meter of air (µg/m3). The ratio of the given gases is given (µg/m3) so, it will be converted into ppm. Depending on this standard, the measured gas in ppm and ppb should not exceed the range as specified in the table below [9] [26].

Table 4-2 shows the ambient concentration range of different gases based on WHO (world health organization).

**Table 4-2: Permitted standard concentration level**

| Substance | Permitted ambient concentration | Permitted industrial concentration |
|---|---|---|
| **$O_3$** | 0.0473ppm 8h average (100 µg/m3) | 0.1 ppm 8-hour average |
| **CO** | 9 ppm 8h average and 35 ppm 1hour average | 25 ppm 8-hour average |
| **$CO_2$** | 5000 ppm 8-hour average | - |
| **Dust** | 10 (µg/m3) annual average and 25 (µg/m3) 8-hour | - |
| **LPG** | 0-1000 ppm (1800 mg/m3) annual average | - |

### 4.2.3 Sensors selection criteria

The main challenge in this project is to identify the type of the sensors based on the kind of application and the way of taking reading from the sensors (outdoor, indoor) application. So, the type of sensors installation will be deployed requires the measurement of concentrations of Gases in air for a specific range [3]. The choice of sensor is critical, since our experimental need accurate result to investigate the status of the weather condition in the state. These results will indicate different percentage of weather contamination in a specific location.

Gas sensors will be compared based on some criteria to get best options using the best optimization methods, in addition the sensors were filtered accordingly to specific requirement

1. Power consumption: since energy consumption is an important restriction, so, total amount of energy required to be calculated for different kind of configurations. It should be noted that a measurement requires initial preheat and stable sensor readings, which must be allowed in the calculations. Also, the reading must be taken from the sensor in the high level of consumption, taking readings during different weather condition and different level of contamination.
2. Size: small sensor size for embedding in the physical node (ability to include sensors in the plain body)
3. Calibration: no recalibration while the airplanes are running.
4. Measurement accuracy: must be in the possible range and meet world health organization's standards.
5. Low-cost: the sensor price is important to get efficient system with low cost and this is engineering management.
6. Sensitivity: the change of measured signal output due to the change of analytic concentration unit, i.e., the slope of a calibration graph in the data sheet. This parameter is affected with the detection limit range.
7. Selectivity (cross sensitivity): this option determines whether a sensor can selectively respond to a group of analytic (chemicals gases) or even to a single analytic.
8. Stability: the ability of a sensor to provide numerous results for a specific period. This includes keeping the state of the sensitivity, selectivity, response, and recovery time in a good condition.
9. Linearity: the relative variation of an experimentally determined calibration graph compared to ideal straight line, more linear easier to deal with(calibration).
10. Resolution: the lowest difference in concentration that can be distinguished by sensor.
11. Response time: the required time for sensor to respond to concentration change from zero to a certain value.
12. Recovery time: the time for the sensor signal to return to its initial value after a change in the concentration value from a certain value to zero.
13. Working temperature: is usually the temperature that corresponds to maximum sensitivity.
14. Offset: represent the difference between the actual output value and the specified output value under various kinds of conditions.

**Table 4-3: $CO_2$ sensor comparison table**

| **Sensor type** | COZIR Ambient | CO2 Engine K30 FR | OS IAQ Sensor VS | Telaire 6613 | MG811 | MQ135 |
|---|---|---|---|---|---|---|
| **Measurement range** | Ambient–5,000 ppm | Ambient–5,000 ppm | Ambient–2,000 ppm | Ambient–2,000 ppm | 350-10000ppm | 10-1000 ppm |
| **Accuracy** | ±50 ppm +/− 3% of reading | ±70 ppm +/− 5% of reading | ±<50 ppm | ±30 ppm +/− 5% of reading | - | - |
| **Power supply** | 3.2 to 5 V | 4.5 to 14 V | 16 to 38 V | 5 V | 6.0±0.1 V | 5v |
| **Power consumption** | 3.3mW | 315mW | 500mW | 165mW | 1200mW | 900 mw |
| **Operating temperature** | -30 to 70°C | 0 to 50 °C | 10 to 40°C | -40 to70 °C | -20 to 50°C | -10-50 |
| **Warmup time** | 1.2 sec | 60 sec | <60 sec | <120 sec | - | 48 hour |
| **Sampling time** | 0.5 sec | 2 seconds @ 0,5 l/min tube gas flow | - | 4 sec | 7 sec | 0.25 second |
| **Dimensions (mm)** | 42x17x14 | 57x51x14 | - | 57x 34x 15 | 37.5x50.8 | 23.3 x30 x18 |
| **Availability** | × | × | × | × | Available | Available |
| **Cost** | $109 | $35 | $95.00 | $144 | $47 | 8$ |

From the above comparison, MQ135 sensor seems the practical option from the rest due to its availability and the high cost of other sensors. Also, the MQ135 sensor does not need an external source for pre-heating compared with MG811. However, it has serious drawbacks, such as the detection range of gases is quite big. Moreover, this sensor needs proper calibration in order to get more accurate results which could be a part in future work since it need special kit for calibration.

**Table 4-4: CO sensors comparison table**

| **Sensor type** | 3SP_CO_1000 Package 110-102 | MQ-2 | AS-MLV-P2 | NE-4-Nemoto | Euro-sensor Eco-sure |
|---|---|---|---|---|---|
| **Measurement range** | 0 to 1,000 ppm | 200 ppm-10000ppm | 0ppm-20ppm | 0 – 1000 ppm | 0-200 ppm |
| **Accuracy** | 4.75 ± 2.75 ppm | High sensitivity | High sensitivity | 75 +/- 15 ppm | 200nA+/-33 |
| **Power supply** | 0 to 5 mV | 5V±0.1 | 3.0 V | 5V | 5V |
| **Power consumption** | 10 to 50 uW | About 900mW | 50 mW | 0.1 W | 500 mw |
| **Operating temperature** | -30 to 55 °C | -20°C-50°C | 0°C to 50°C | -20°C to +50°C | 10- 30 °C |
| **Warmup time** | 2 hour | Not less than 24 hours | - | No heater | No heater |
| **Sampling time** | 15 sec | 0.25 sec | < 10sec | < 30 sec | < 30 sec |
| **Dimensions(mm) & weight** | 20.8 x 20.8x7.69<br>- | 35x 28 x23<br>- | 9.1x9.1x4.5<br>- | -<br>5 g | -<br>5 g |
| **Availability** | × | ✓ | × | × | × |
| **Cost** | $20 | $9 | $400 | $150 | $30 |

From the comparison table for CO sensors above, AS-MLV-P2 sensor has very limited range 0ppm to 20ppm which is appropriate for indoor applications such as kitchen and house application. Moreover, it has the highest cost of the three sensors. After excluded AS-MLV-P2 now three options are remaining MQ-2 and 3SP_CO_1000Package 100-102, MQ-2 has wider range from 200ppm to 10000ppm but this sensor will not give any range before that. Also, MQ-2 sensor available in the campus with cheapest price 9$ and need 24 hour of pre-heat. Moreover, this sensor measure several gases based on the calibration process example (LPG gas) with a different RS value. Otherwise, 3SP_CO_1000Package 100-102 has less measurement range 0 ppm to 1000ppm and not available. In addition, it needs a long time power-on stabilization around 2 hours, the last sensor has a good range and it has suitable price but it has a temperature problem (low range from 10 to 30), Based on this result the MQ2 sensor could be the best option for the CO sensor.

**Table 4-5: $O_3$ sensors comparison table**

| Sensor type | MQ131 model | MS2610 model | 3SP_O2_20 | Mica 2610 |
|---|---|---|---|---|
| **Measurement range** | 10-2000ppb | 0-10000 ppb | 0 to 20 ppm | 10-1000 ppb |
| **Accuracy** | - | - | < 20 ppb | 50 -70 ppb |
| **Power supply** | 5.0 V | 5.0 V | 5.0 V | 5 V |
| **Power consumption** | <900 mW | <650mW | 10 to 50 uW | 100 mw |
| **Operating temperature in** | - | 0-50 °C | -30 to 50 °C | -40 to 70 °C |
| **Warmup time** | Over 48 hours | - | 2 hour | - |
| **Sampling time** | - | >30sec | < 15 sec | 30 sec |
| **Dimensions(mm) & weight** | 32x22x30<br>0.025 kg | 40x35x18<br>15g | 20x20x3<br>- | 13.9x9.4<br>- |
| **Availability** | × | × | × | × |
| **Cost** | 23$ | 37$ | 20$ | 40$ |

From comparison table of $O_3$ there are four sensors. First, MQ131 model has a good measurement range, but it is consuming 950mW and take around 48 hours to warmup one time only. The second model MS2610 has a good measurement range also lower power consumption 650mW. Third sensor 3SP_O2_20 P Package 110-401 have a good range also and have low power consumption but it has high power on stabilization. Mica 2610 have lower power consumption and less space dimension but it need external circuit which is may not accurate. However, MS2610 model, has a higher price but it has a good range and low power consumption, but it is not available in the campus. So, the best and the applicable one is the MQ131 which will be used.

**Table 4-6: Temperature sensor comparison table**

| **Sensor** | AM2302/DHT22 | DS18B20 | LM35DT/NOPB |
|---|---|---|---|
| **Operating range** | Humidity: 0-100%RH<br>Temperature: -40 ~ +80 ºC | -55 ~ +125 ºC | 0 to 100 ºC |
| **Accuracy** | Humidity: ±2%RH<br>Temperature: ±0.5ºC | ±0.5ºC | ±0.75ºC |
| **Power supply** | 3.3-5.5V | 3.3V to 5V | 4 V to 30 V |
| **Power consumption** | <9mW | <7.5mW | 15mW |
| **Response Time** | 2 sec | 1.5 sec | N.D |
| **Dimensions(mm) & weight** | 27x58.75x13.3<br>8.5g | 22x32<br>5g | 10.16x 4.7x8.89<br>2.04 g |
| **Availability** | ✓ | ✓ | ✓ |
| **cost** | $15 | $7.5 | $2.32 |

AM2302 is wired module of DH22, it utilizes a capacitive humidity sensor and a thermistor in order to measure both humidity and temperature of surrounding air. The sensor seems to be the best option in terms of high precision, stability, power consumption and fully automated calibration. In addition, AM2302 comes with a single wire digital interface unlike LM35DT that has analog data output, so it does not require extra components. Furthermore, this sensor is chosen since it measures humidity and temperature at the same time, which reduces the used sensors.

**Table 4-7: Dust sensors comparison table**

| **Sensor** | PPD42NS | SM-PWM-01C |
|---|---|---|
| **Operating range** | 0~45°C | -10~45°C |
| **Detectable particle size** | >1μm | >1μm |
| **Supply voltage** | 5V | 5V |
| **Power consumption** | 90mA | 90mA |
| **Time for stabilization** | 60sec | 90sec |
| **Dimensions(mm) & weight** | 59 × 45 × 22<br>24g | 59 × 46 × 18<br>20g |
| **Availability** | ✓ | × |
| **cost** | $21 | $14.8 |

These sensors are used in order to measure the air quality in an environment by measuring the dust concentration especially for outdoor applications. The Particulate Matter level (PM level) in the air is measured by counting the Low Pulse Occupancy time (LPO time) in given time unit. LPO time is proportional to PM concentration. Both sensors can provide reliable data for air purifier systems; they are responsive to PM of diameter 1μm. From the table above both sensors have similar specifications. However, SM-PWM-01C is better in term of operating range and it is lighter than PPD42NS. Although, PPD42NS will be used in this project because it is available in the campus and it has less time for stabilization.

In short, ideal sensor would appropriate high sensitivity, dynamic range, selectivity, and stability; low detection limit; good linearity; small hysteresis and response time; and long life cycle [27]. Researchers usually looking for approaching only some of these ideal characteristics and ignore the rest.

The task of making an ideal sensor for some gases is extremely difficult, if possible. Therefore, real applications usually do not require sensors with all perfect characteristics at once. For instance, monitoring the concentration of gases in industrial areas does not need high detection limit, on the other hand, the response time should be in seconds or less. While in case of monitoring the environment, detection limit is important, and much higher than the concentrations of pollutants, which normally slow.

In this project, sensors should have fast response time; fast recovery period, in addition to make sure of the suitable range of ppm reading and the other preference.

Calibrating the sensors, especially the gas sensors need to understand the way of working for every sensor, and every agent that might affect by it. The gas sensors that was selected are not calibrated to all the sensor (cross sensitivity only for some sensors, and based on the expected measurement the manufacture can afford extra calibration based on client request). Also, this sensor should be calibrated in presence of high level of contamination ppm gas, to maintain the largest quantity that can be handled by the sensor's.

### 4.2.4 Selected sensors characteristics

The table below shows some information regarding the sensor that will be used in this project, the required number of pin, pin classifications, and the type of the sensor in order to help us to selected the suitable microcontroller that satisfy this requirement. So, for the sensors part only, the micro-controller should be satisfying the sensor requirement, at least capable of providing 4 terminals of digital input and 5 analog pins in the sensors case only beside that the other component from different part from the airplane.

**Table 4-8: Selected sensor specification**

| Sensor Name | Number of pins | Pins classifications | Output signal type | Weight | Price($) |
|---|---|---|---|---|---|
| **MQ131 ($O_3$)** | 4-pins | GND-5v- out | Analog | 25g | $23 |
| **AM2302 Humid. & Temp** | 3-pins | GND-5v- out | Digital | 8.5g | $15 |
| **SM-PWM-01C(Dust)** | 3-pins | GND -output-VCC | Digital | 24g | $21 |
| **MQ2 (CO and LPG)** | 3-pin | GND-5v- out | Analog | 10g | $9 |
| **Mq135 ($CO_2$)** | 3-pin | Ground +VCC+ out | Analog | 10g | $9 |

## 4.3 Weight

UAV performance and functionality are carefully related to plane size, and consequently, small, low-cost aircraft will inherently have payload and wing area restriction. It will has a restricted capacity to hold on-board sensors and probably brief flying time [4]. The Table and pie chart below shows the components weight based on their datasheets.

**Table 4-9: UAV component weight in grams**

| | |
|---|---|
| **Battery** | 819 |
| **Motor** | 470 |
| **Sensors** | 77.5 |
| **Arduino** | 5 |
| **Raspberry Pi** | 45 |
| **Webcam** | 9 |
| **Flight controller + GPS** | 50 |
| **Power Bank** | 280 |
| **Others** | 100 |
| **Total** | 1855.5 |

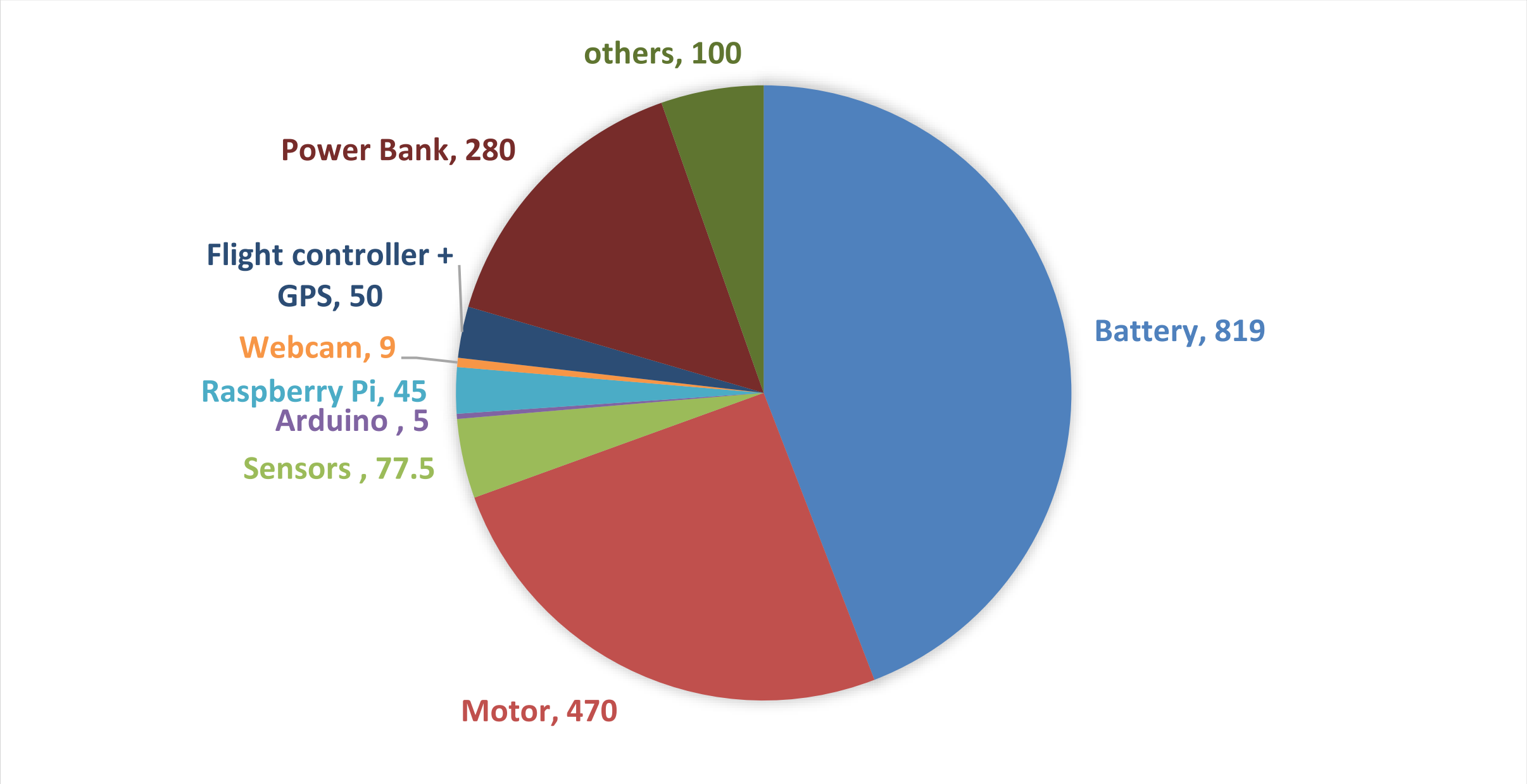

**Figure 4-7: Pie chart of total weight in grams**

## 4.4 Project Cost

The table 4-16 illustrates briefly the net cost of the project.

**Table 4-10: Approximation of total cost**

| | |
|---|---|
| Airplane | $2600 |
| Battery | $600 |
| Sensors | $400 |
| Arduino | $16 |
| Raspberry Pi | $82 |
| Camera | $60 |
| 4G USB | $160 |
| Flight controller | $400 |
| GPS | $90 |
| Power bank | $98 |
| **TOTAL** | **$4506** |

# Chapter 5 PRACTICAL IMPLEMENTATION AND RESULTS

This chapter presents the practical implementation of the sensors, communication and autopilot. In addition, practical results are discussed illustrated.

## 5.1 Sensors

### 5.1.1 Sensors performance in different speeds

The circuit in Figure 5-1 has been implemented and tested over different speed by fixing it on a car and taking several readings for each speed in order to notice the effect of the airflow on sensor readings.

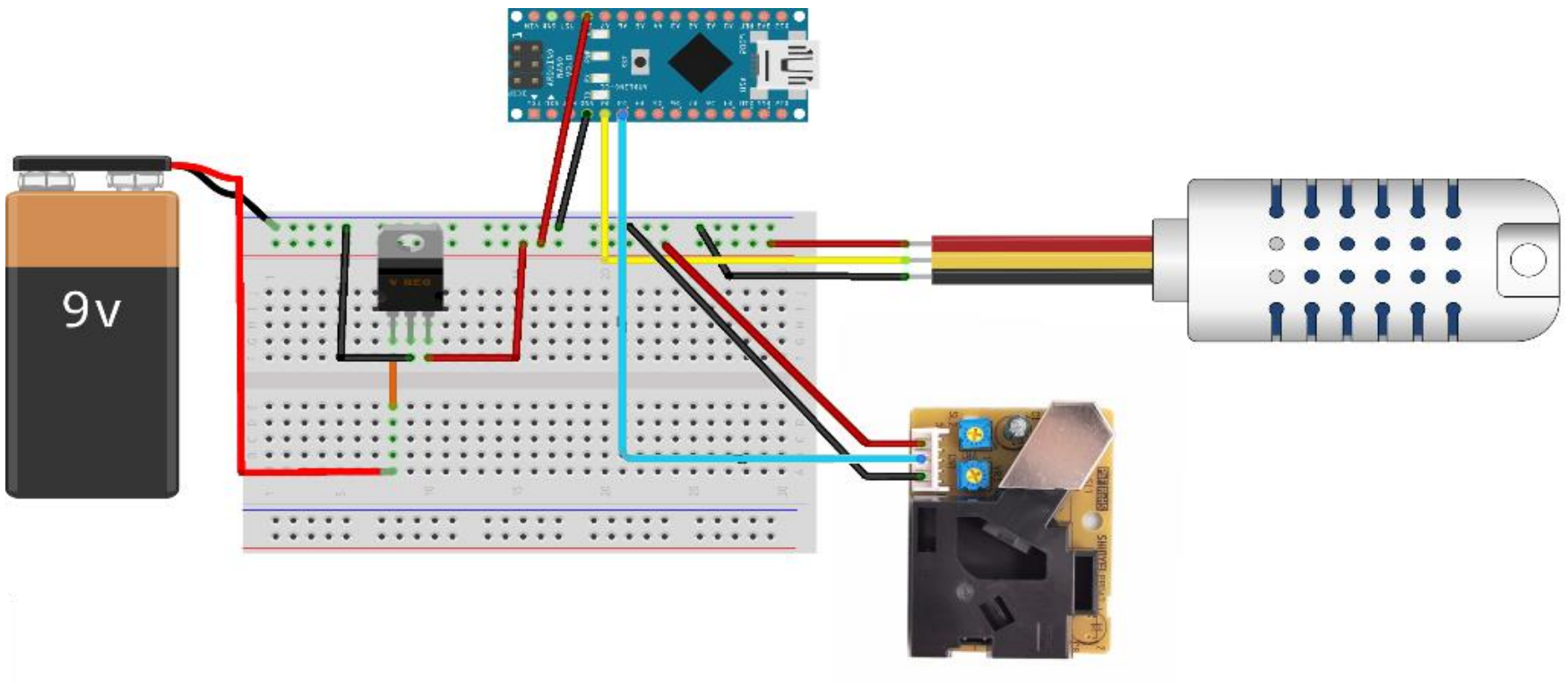

**Figure 5-1: Implemented circuit of Temp, Humidity and Dust sensors with Arduino**

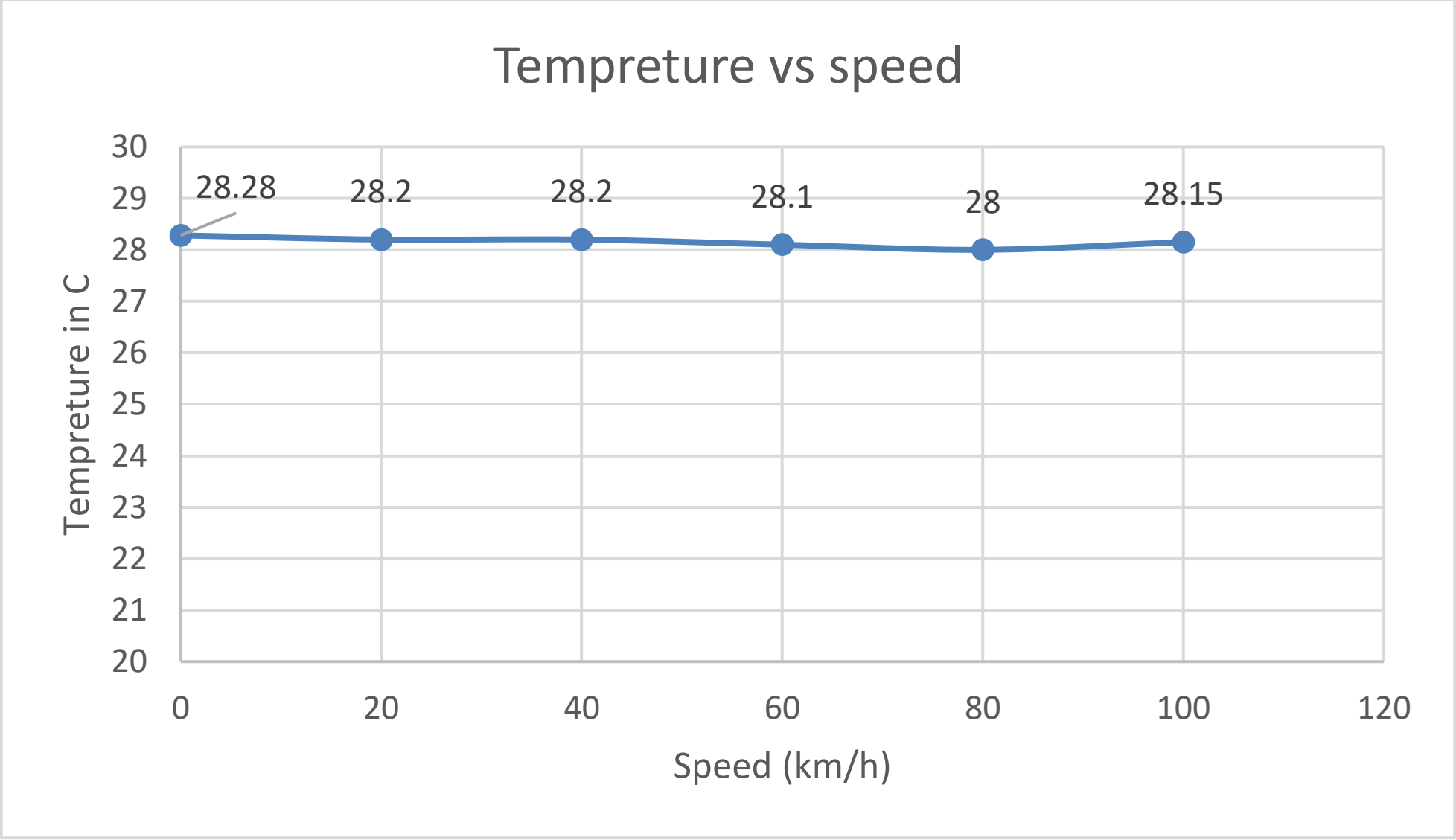

**Figure 5-2: Temperature Vs Speed**

The Figure 5-2 proved that the airflow has insignificant impact on the temperature sensor AM2302. Similarly, the effect of the airflow on humidity measurements is slight also as shown in Figure 5-3.

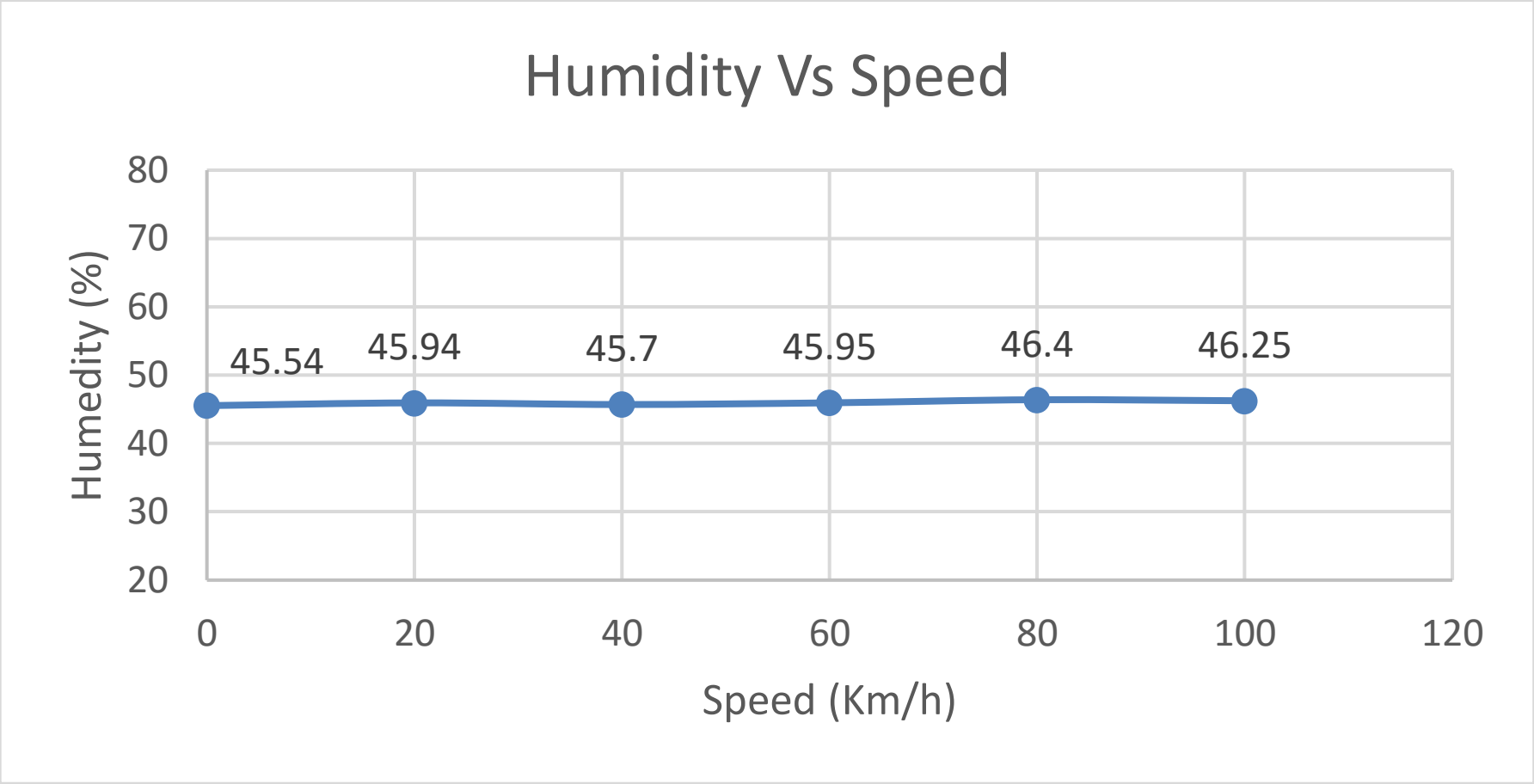

**Figure 5-3: Humidity Vs Speed**

It is clearly noticed from Figure 5-4 that by increasing the speed the value of dust concentration is also increasing which means that the airflow has major impact on sensor readings.

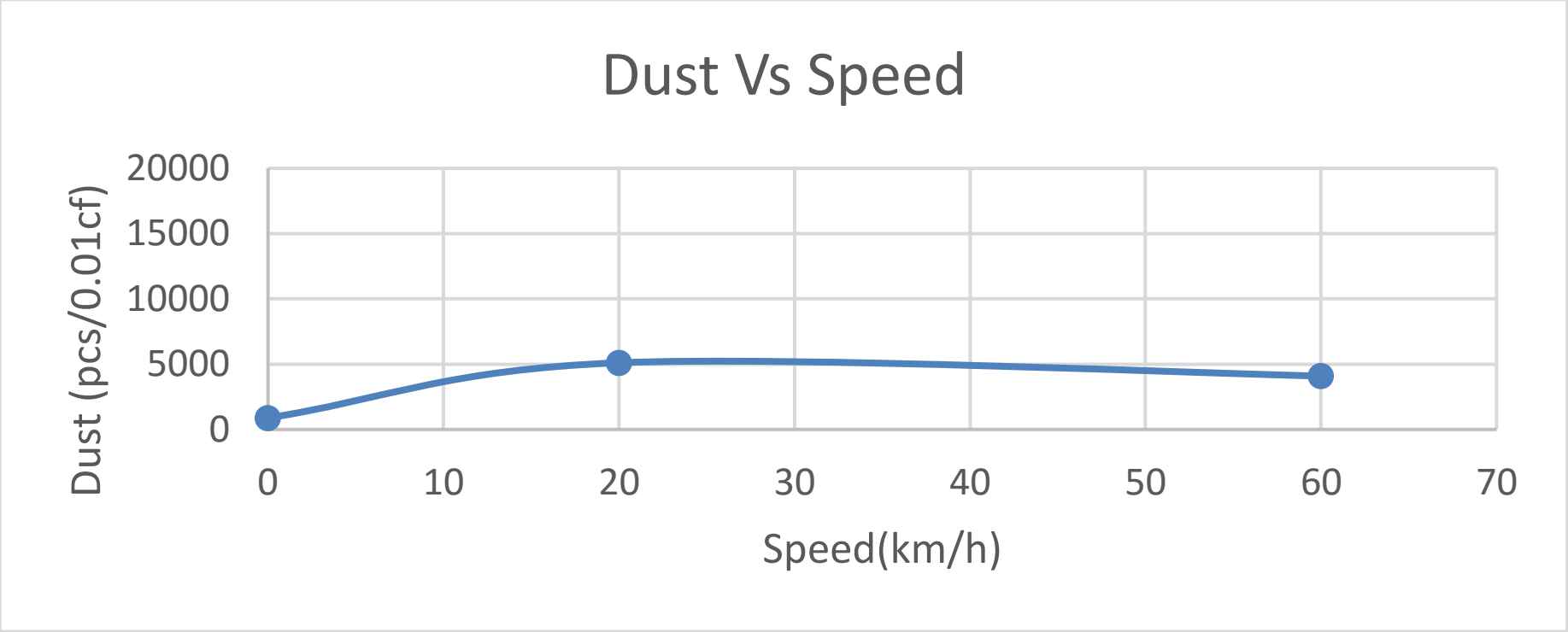

**Figure 5-4: Dust Vs Speed**

### 5.1.2 Sensors preheating and testing

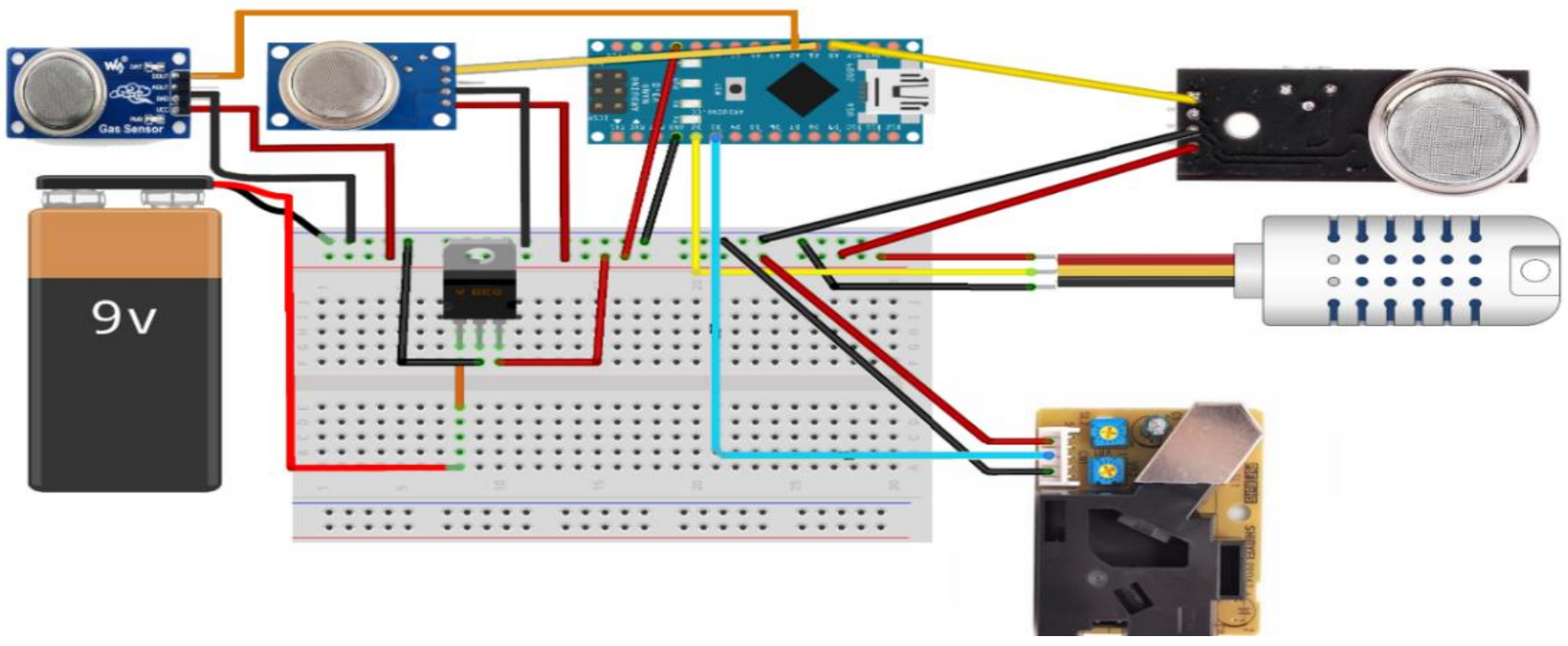

**Figure 5-5: Implemented circuit of Arduino with all sensors**

The circuit in Figure 5-5 implemented and kept on connected for at least 24 hours in order to preheat sensors to make the values more consistent since when sensor gets warm that means it can work normally and sensitively at that time. It's one time process, so no need to repeat it later.

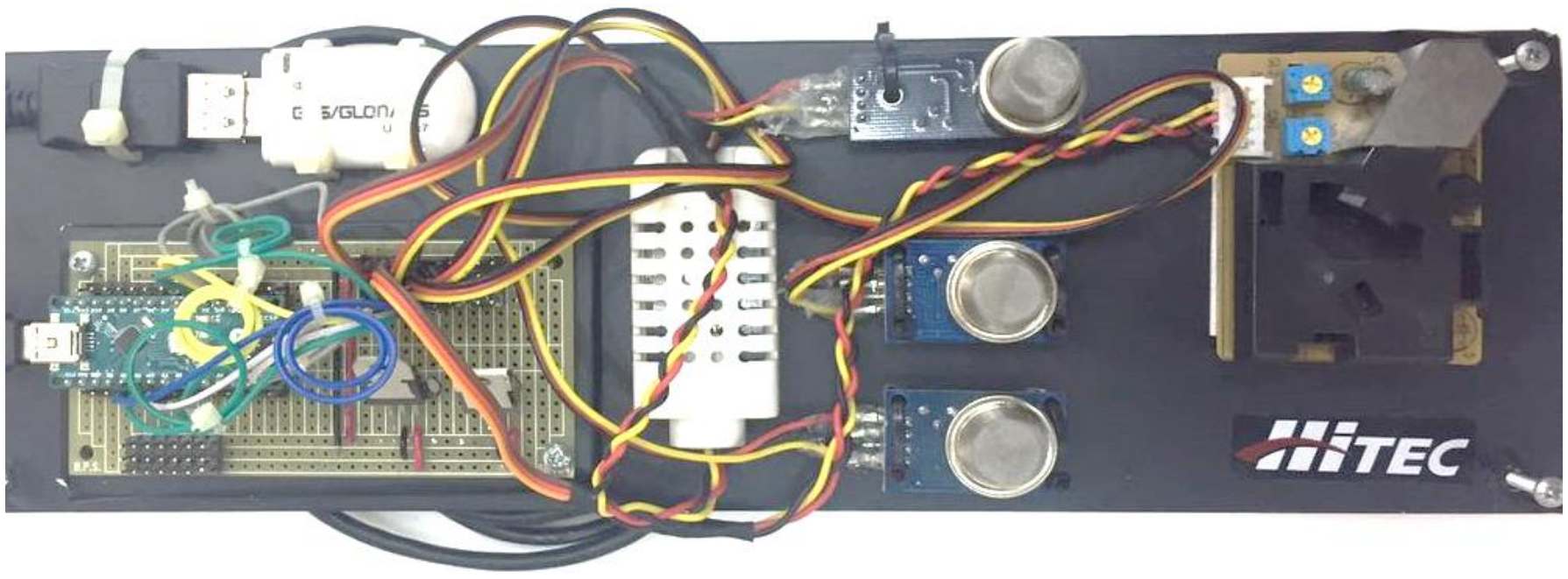

**Figure 5-6: Sensors practical implementation**

First, the result in room ambient is 0 ppm as shown in figure 5-7 which meet the ASHRAE standards of CO concentration in normal fresh air. Then, we placed a lighter close to the MQ-2 gas sensor, and pressed the switch to release gasses in purpose of noticing the effect on sensors reading.

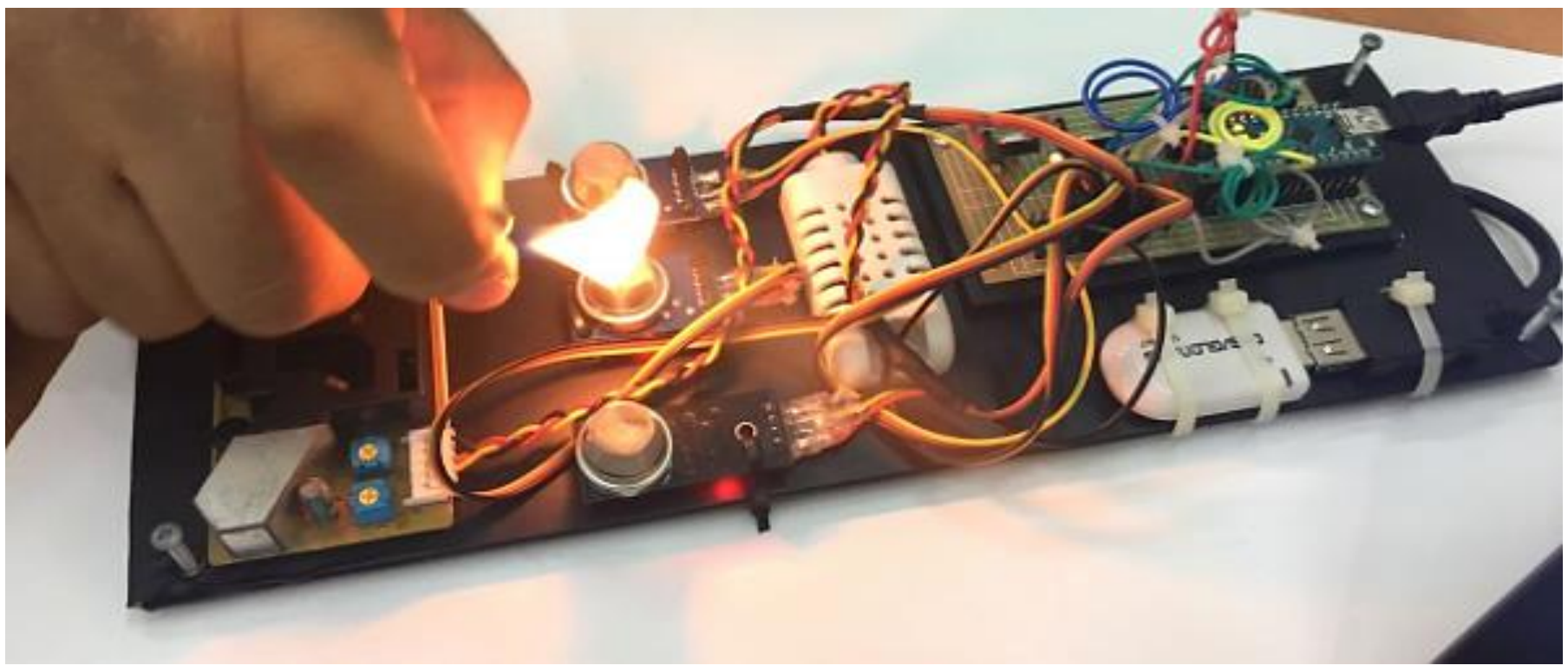

**Figure 5-7: Placed a lighter gas close to the MQ-2**

Figure 5-8 shows that the effect of releasing gasses from the lighter.

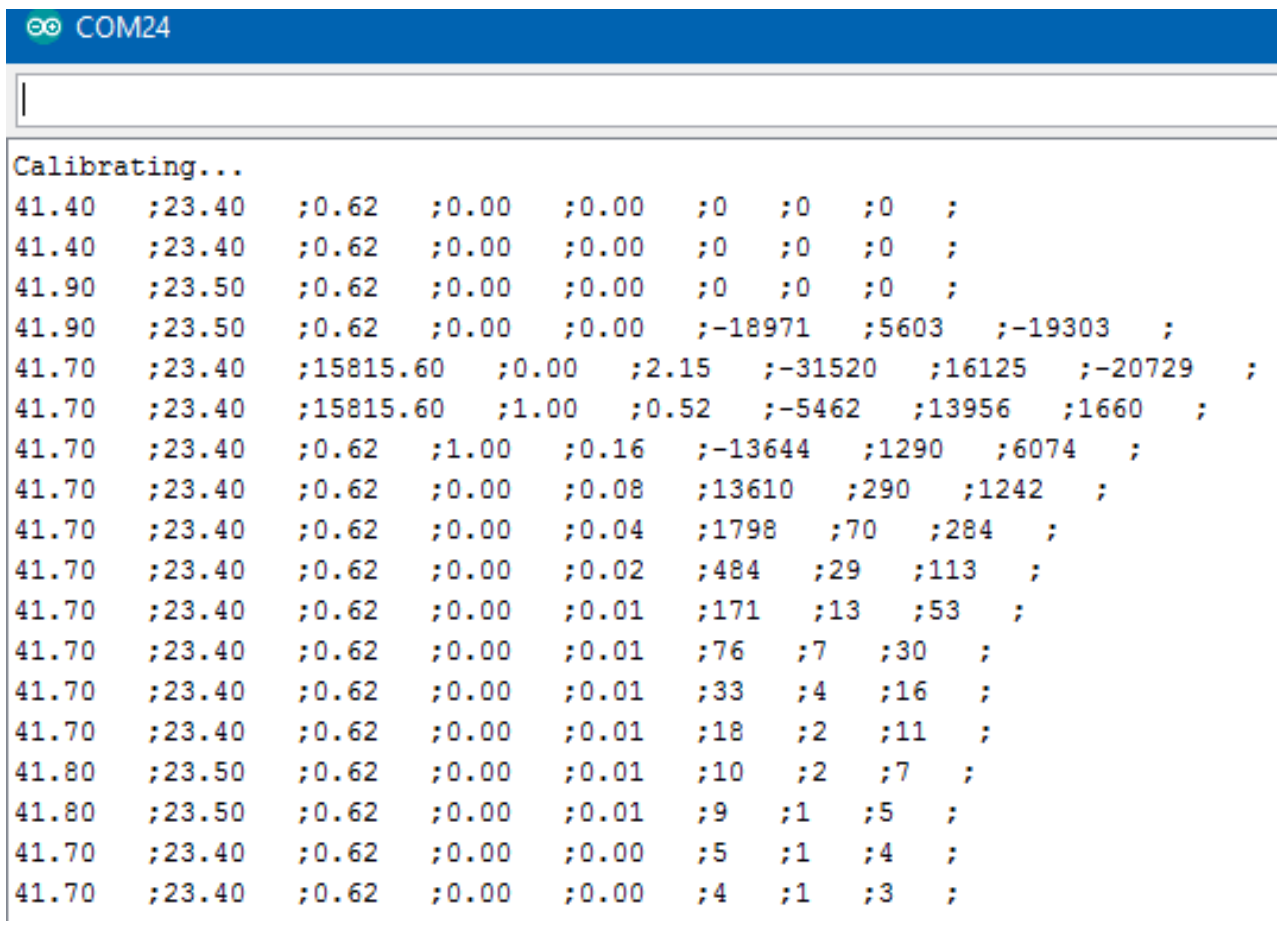

```
COM24

Calibrating...
41.40   ;23.40   ;0.62   ;0.00   ;0.00   ;0   ;0   ;0   ;
41.40   ;23.40   ;0.62   ;0.00   ;0.00   ;0   ;0   ;0   ;
41.90   ;23.50   ;0.62   ;0.00   ;0.00   ;0   ;0   ;0   ;
41.90   ;23.50   ;0.62   ;0.00   ;0.00   ;-18971   ;5603   ;-19303   ;
41.70   ;23.40   ;15815.60   ;0.00   ;2.15   ;-31520   ;16125   ;-20729   ;
41.70   ;23.40   ;15815.60   ;1.00   ;0.52   ;-5462   ;13956   ;1660   ;
41.70   ;23.40   ;0.62   ;1.00   ;0.16   ;-13644   ;1290   ;6074   ;
41.70   ;23.40   ;0.62   ;0.00   ;0.08   ;13610   ;290   ;1242   ;
41.70   ;23.40   ;0.62   ;0.00   ;0.04   ;1798   ;70   ;284   ;
41.70   ;23.40   ;0.62   ;0.00   ;0.02   ;484   ;29   ;113   ;
41.70   ;23.40   ;0.62   ;0.00   ;0.01   ;171   ;13   ;53   ;
41.70   ;23.40   ;0.62   ;0.00   ;0.01   ;76   ;7   ;30   ;
41.70   ;23.40   ;0.62   ;0.00   ;0.01   ;33   ;4   ;16   ;
41.70   ;23.40   ;0.62   ;0.00   ;0.01   ;18   ;2   ;11   ;
41.80   ;23.50   ;0.62   ;0.00   ;0.01   ;10   ;2   ;7   ;
41.80   ;23.50   ;0.62   ;0.00   ;0.01   ;9   ;1   ;5   ;
41.70   ;23.40   ;0.62   ;0.00   ;0.00   ;5   ;1   ;4   ;
41.70   ;23.40   ;0.62   ;0.00   ;0.00   ;4   ;1   ;3   ;
```

**Figure 5-8: Sensors readings**

## 5.2 *Communication setup*

The UAV module consists of two main components which are data collecting component and navigation component which are controlled via the main controller.

### 5.2.1 Data collecting component

To measure data and collect it, a circuit consisting of an Arduino and a set of sensors was implemented. These sensors take different reading on a set of intervals and sent the data via the USB cable to the raspberry pi. The reading from the sensor was calibrated previously and the data will be sent to the Arduino.

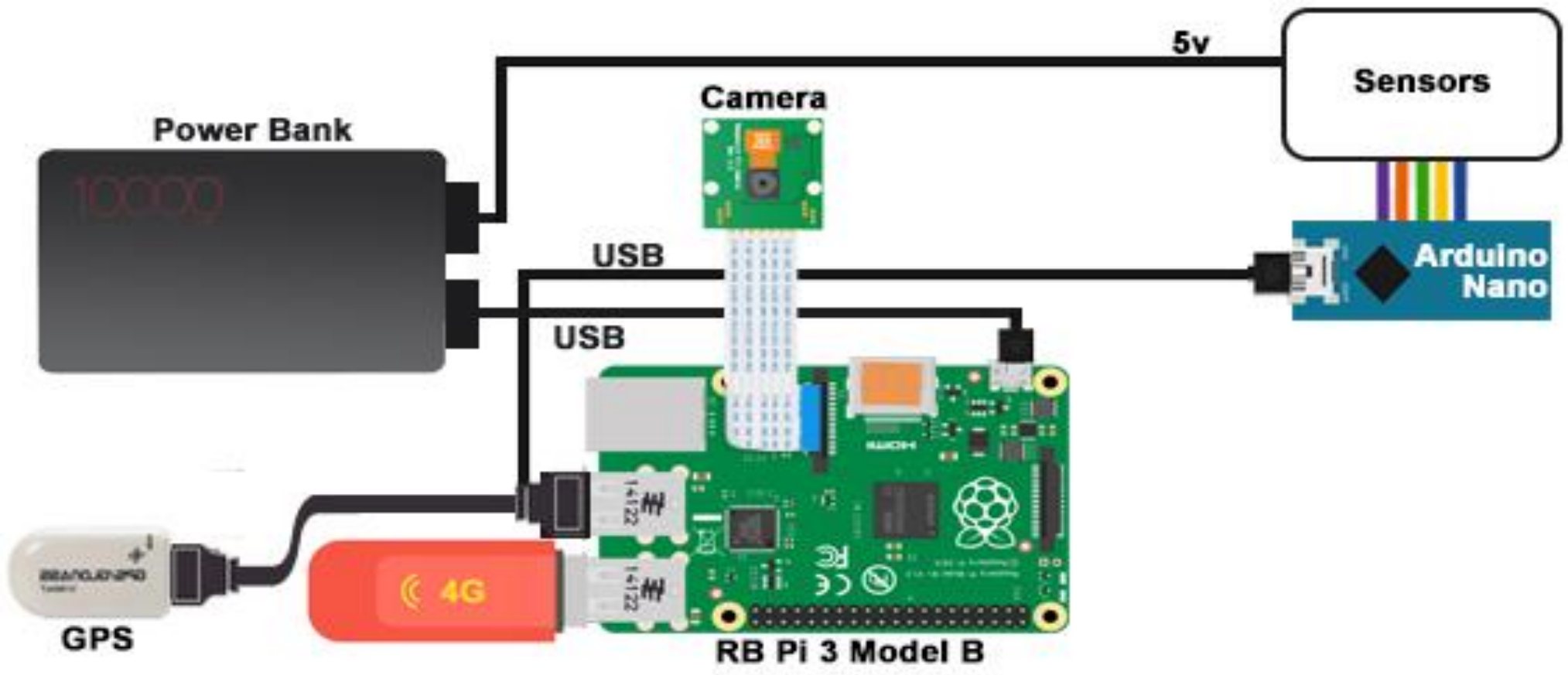

**Figure 5-9: Wiring diagram of data collecting and transmitting**

As shown above, we are using a power bank with two output ports to provide a constant 5V, one port is used to power on the Raspberry Pi and the other to power on the sensors. This solution is better than using battery combined to voltage regulator because of voltage regulator heat issue.

### 5.2.2 Auto-pilot navigation and data transmission

Due to plane crash on the first flight, it was wiser to implement an automatic take-off and landing in conjunction with the manual flight which will protects the UAV from crashing in case of losing of RC signal as well as provide schedule flights.

The idea is to implement an open source Flight Controller (Pixhawk mini) with the Ardupilot firmware installed into it. The Flight Controller acts as the main controller of the navigation. Commands from the RC receiver are wired into it and it will either execute them if it was in Manual flight mode, or ignore them if it is in Auto pilot mode. These commands will be Execution via controlling the main DC motor by sending pulse modulation to the Speed controller (ESC) and controlling the servo motors on the wings and rear of the UAV.

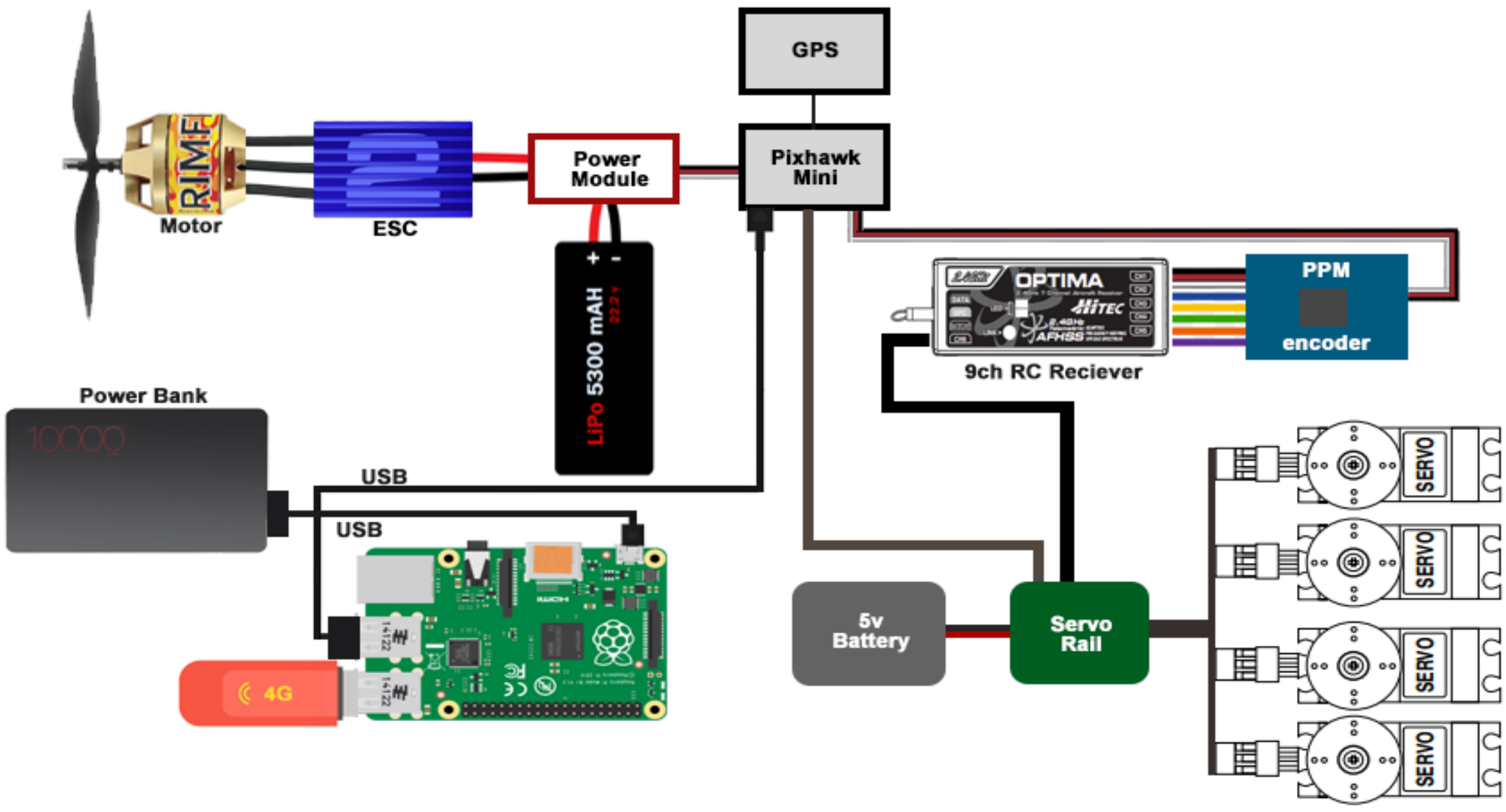

**Figure 5-10: Auto-pilot navigation and data transmission wiring diagram**

The PPM encoder converts the parallel output of the RC receiver into a serial line connected to the Pixhawk mini. The power module ensures the Pixhwak has constant feedback of the current status of the main battery.

The Flight Controller is also connected to the Raspberry pi via a USB connection to broadcast its status as well as receive commands. The main controller is a Raspberry pi. This Raspberry pi is also connected via USB to the GPS to get the current coordinates, and Arduino to get the current measurements of air pollution parameters, as well as the Flight Controller to receive and retransmit the status and give it commands. It is also connected to the camera to receive video stream and transmit server. All these transmissions are done via a LTE modem also connected via a USB to Raspberry pi. In other terms, The Raspberry is the heart of our system and maintaining connection with it is vital.

To accomplish this constant connection, the first step is to setup a reverse SSH tunnel upon Raspberry Pi's bootup with the server. This is done to overcome the obstacle of the 4G modems having no static IP, therefore by setting up the server as a proxy that has a static IP Address, and can forward any requests from a client to the Raspberry pi after a connection is established and authenticated. This mechanism enabled us to SSH to the Raspberry Pi from anywhere if it was powered on with an active internet connection.

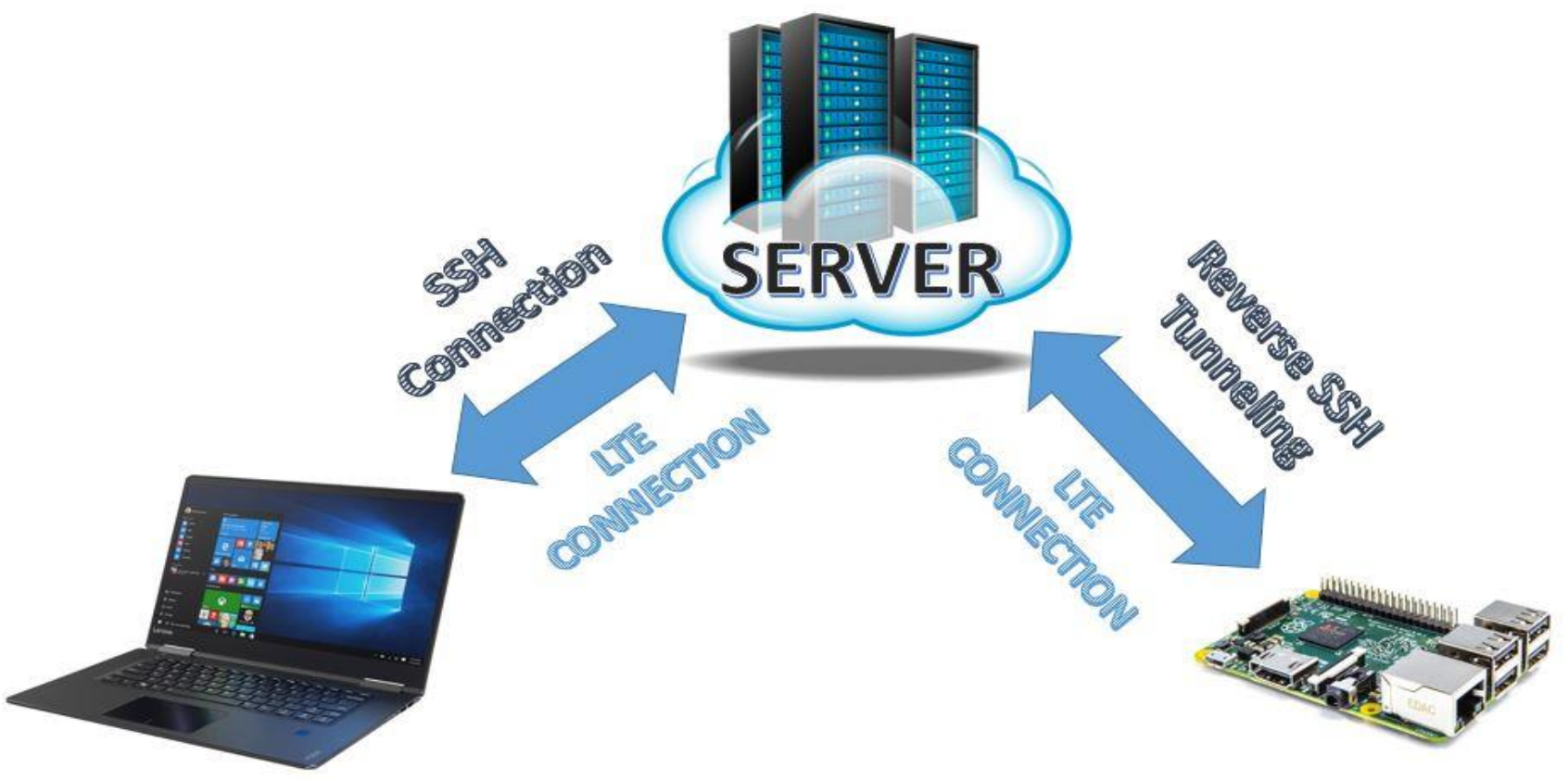

**Figure 5-11: Communication structure**

The next step was to develop a Client/Server application structure. The client resides in the Raspberry Pi to maintain continuously connection the server. After that, the request that come from the server can be either start/stop Sending Data or Start/Stop Sending Video.

The client application has three main threads that work concurrently, the main thread is responsible to maintain connection, and receive requests from the server. The Arduino thread will continuously read data coming from the Arduino and make them available to the main thread. The GPS thread will keep track of the current position and make it available to the main thread.

### 5.2.3 Safety

The Flight Controller is powered via a battery that powers the servos on the UAV, as well as the main DC motor battery and the Raspberry Pi. This is done to make sure that the Flight Controller is constantly aware of the status of each of these power sources and can give proper waring and act in case of powers loss of one or two sources of power. In addition to, a manual override switch is configured in the RC transmitter which enabled manual override of commands in case anything goes wrong with the autonomous system, And the Autopilot is configured to attempt to return to base and hover around it in case of communication loss.

### 5.2.4 Data viewing module

The Data viewing module consists of a website and an android application, the website features Tables, Graphs and Google map that shows current data stored, and can be filtered geographically or based on time and date. While the android application features getting closest stored measurement to alert the user of the status of air pollution in his area of a general google map with colored markers that show all measurements in the specified period.

## 5.3 Final implementation

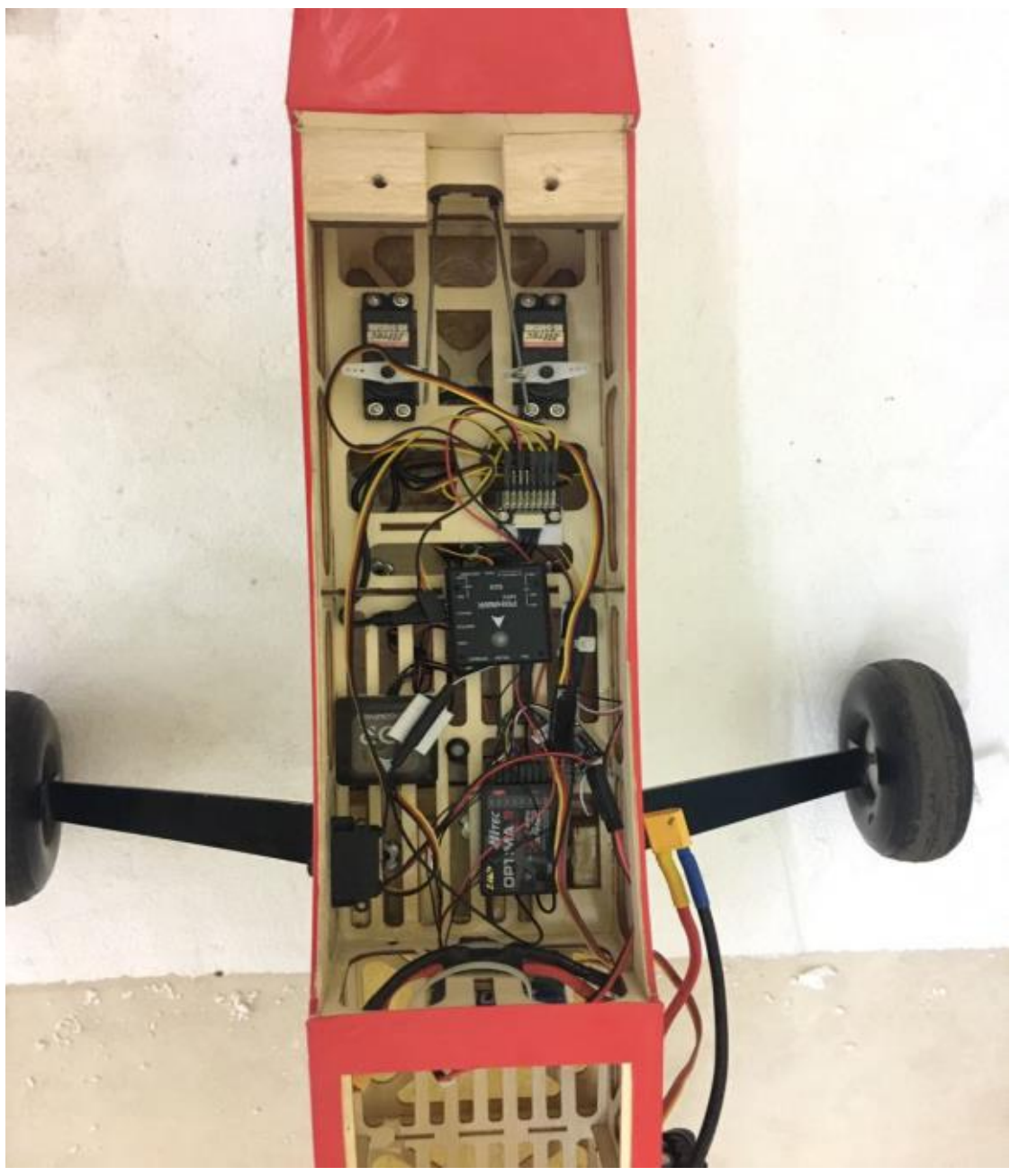

**Figure 5-13: Assembled Navigation system**

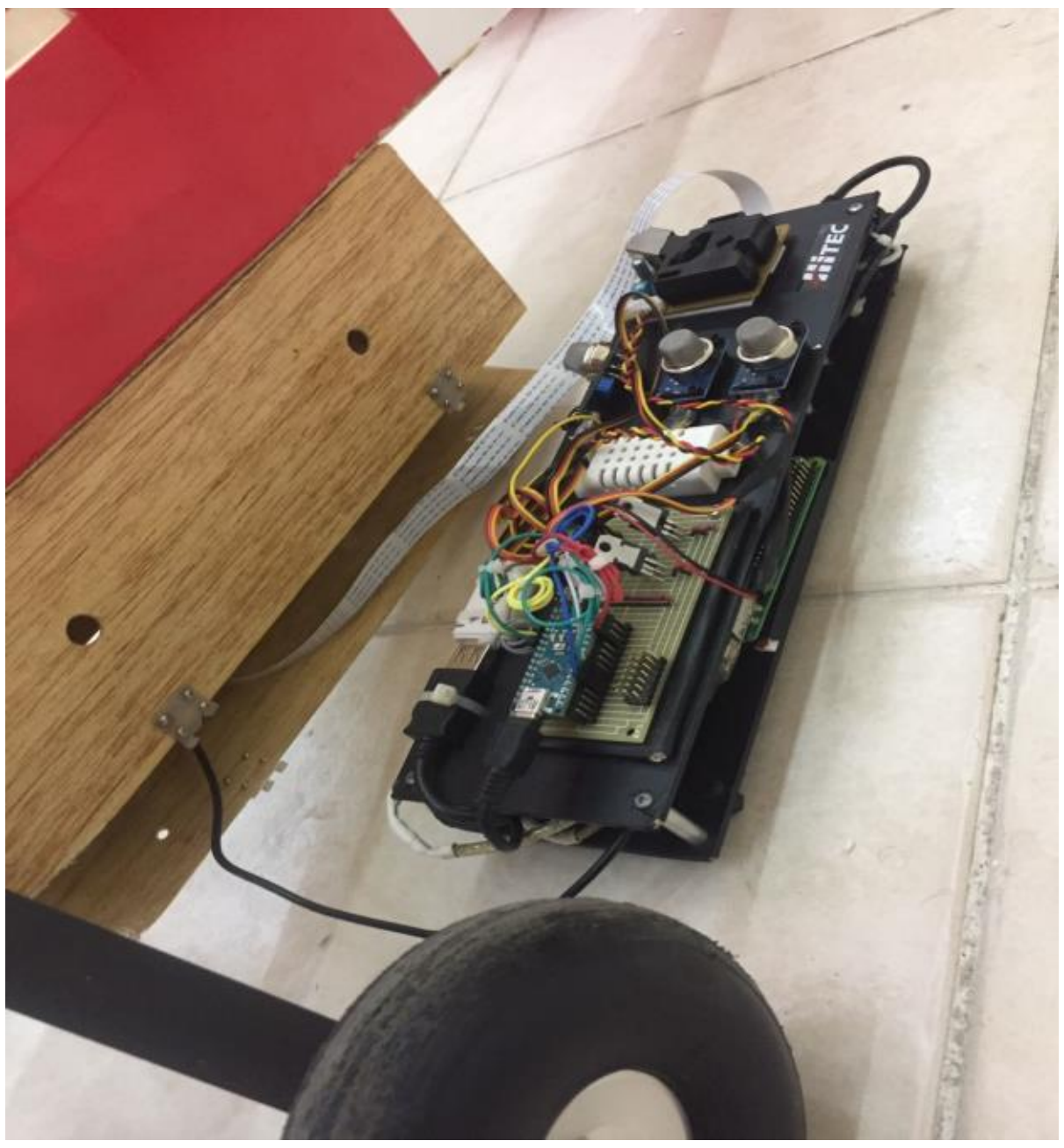

**Figure 5-13: Assembled Data collection module**

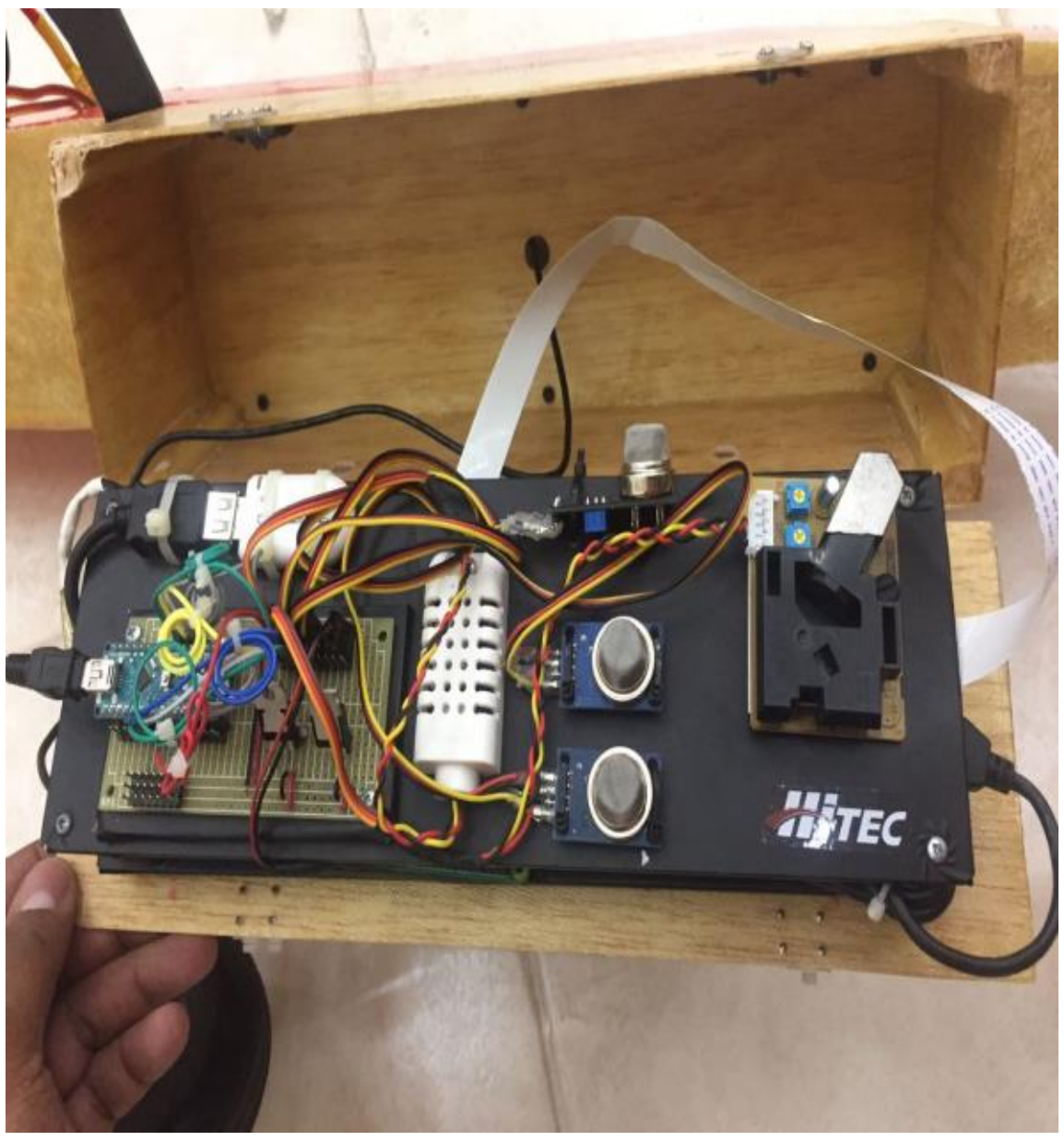

**Figure 5-15: Data collecting module inside the chamber in the UAV**

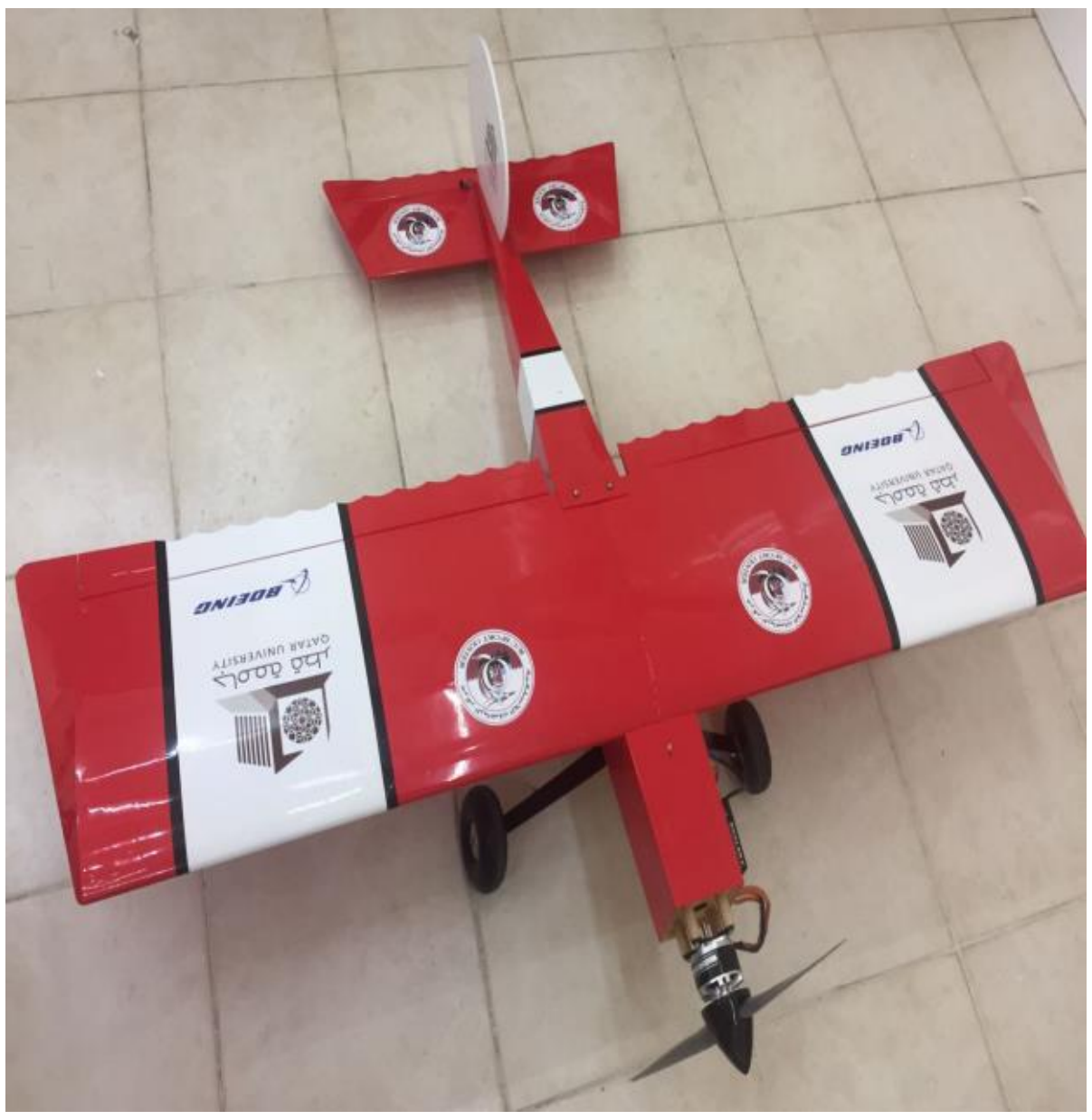

**Figure 5-15: Finalized UAV module**

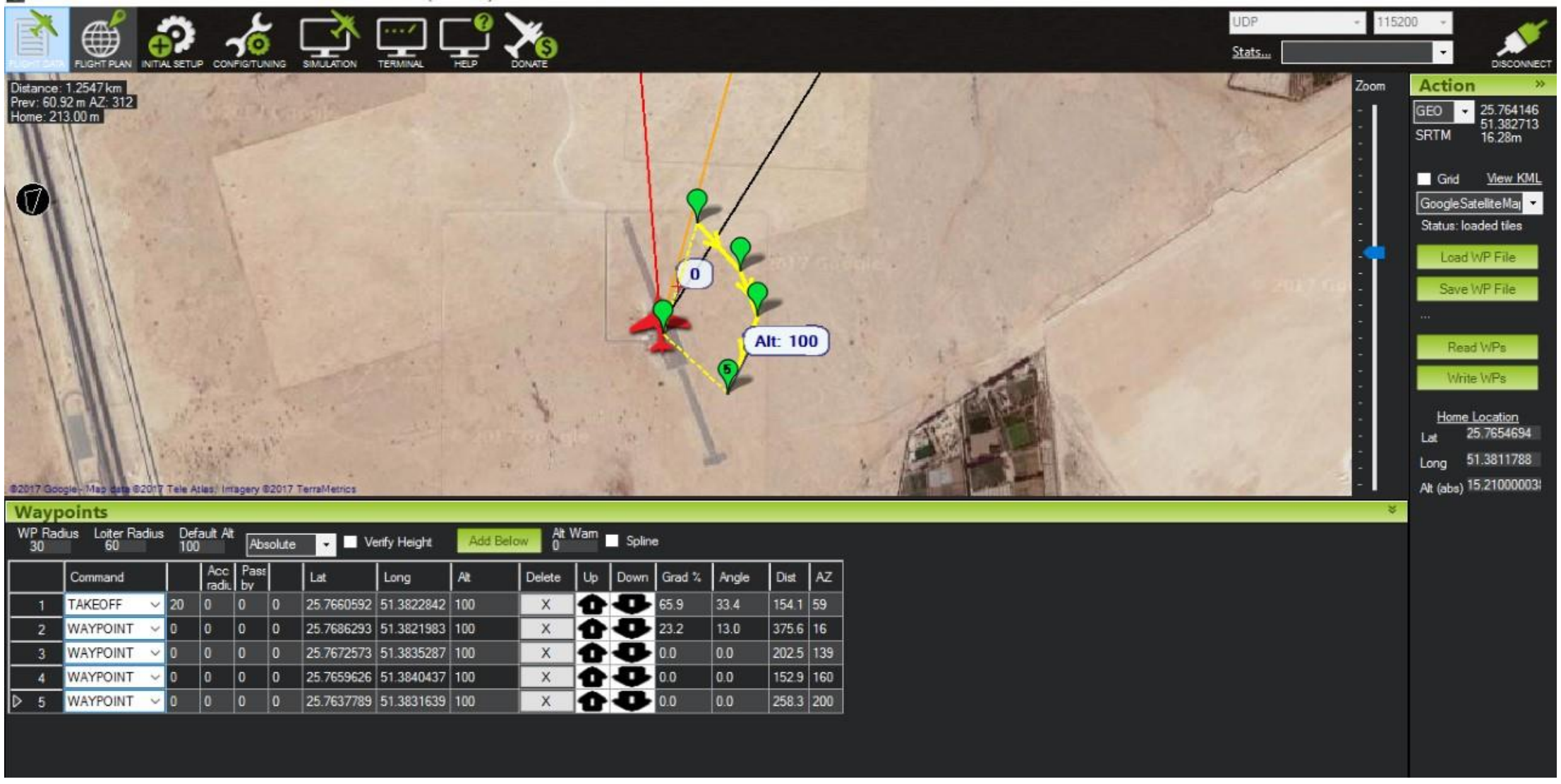

**Figure 5-16: Mission planner**

## 5.4 *Testing and result*

The project involves a UAV which makes it especially fault intolerant. Therefore, each component had to be thoroughly and rigorously tested individually and after integrating with the other components of the system. In general, the component can be classified as the following categories:

### 5.4.1 Payload test

The first test conducted by the team was the pay load test. The objective of this test is to examine if the plane can carry the weight of the electronic component. The test took place in Qatar RC center. A weight of 1 KG (which is above the weight of the electronic component) was placed in the circuit cabin then with the help of a member of the RC club the plane could sustain a constant fight.

### 5.4.2 Navigation

This was the trickiest part of the project. On the first test flight a loss of communication caused an immediate crash landing, causing the destruction of all the equipment. This failure has pushed the team to work tirelessly to integrate a flight controller in the system. Having done so, the team started testing the auto pilot system's reactions to sudden loss of altitude and sudden changes in orientation. In all cases the autopilot responded swiftly and attempted to restore both its altitude and orientation, these tests were conducted on the ground before flight. Due to its riskiness, this test was divided to three stages.

- **Stage 1:**
  - Setting the mode to Flight Assistant mode
  - Holding the aircraft up while it is power on
  - Rotating the aircraft
  - Measuring response time from the Flight Controller to stabilize the aircraft

**Results:**

Response time was always less than one second. After integrating the flight controller into our navigation system as a failsafe to protect the UAV from crashing, we started testing manual flights, in all tests the UAV behaved normally and remained stable during flight. Having made sure that the flight is configured properly to respond to any sudden unwanted change of orientation, we headed to test the autonomous mission capabilities.

We setup test mission involving manual take off, then switch to the autonomous mission to cruise into predefined way points, then return to the base, at that point it will stay circling above the base until the mode is switched back to manual for landing.

- **Stage 2:**
  - Setting the mode to manual.
  - Having a human flight operator perform manual take off.
  - Turn on the Automatic mode once the UAV reaches an altitude of 100 meters.
  - Manually observe the flight path following the mission.
  - Upon mission completion, wait for the UAV to return to base.
  - Having a human operator perform manual landing

**Constraints Observed:**

Setting mission points too close to the point of takeoff will make the UAV attempt to perform steep angle turn midair. It is recommended to keep 100 meters between the take off point and the first point of the autonomous mission.

Setting the last mission point in close proximity of the base, will cause the UAV to circle in narrow radius above the base, thus again resulting in steep turns. It is recommended to keep the last mission point at a distance of at least 200 meters from the base of launch.

**Results:**

In test flights, the UAV successfully passed through all the determined points with an accuracy of 30 meters. Further accuracy can be obtained; however, it will cause extra strain on the flight controller and could cause steep turns.

The next step was setting up auto take off, autonomous flight and manual landing.

- **Stage 3:**
  - Setting the flight controller to automatic mode.
  - The flight controller will automatically start the main motor and take off.
  - Once the UAV reaches an altitude of 100 meters it will head to point 1.
  - Manually observe the flight path is consistent.
  - Upon mission completion, wait for the UAV to return to base.
  - Having a human operator perform manual landing.

**Results:**

The UAV successfully took off, automatically followed the flight plan and returned to base, hovering until manual control was taken and manually landed.

To sum up, in all our test flights, the UAV successfully took off, automatically followed the flight plan and returned to base, hovering until manual control was taken and manually landed.

We had 12 flights in total to test our system. The first flight ended up crashing, while others were successful, maintaining constant connection with both RC transmitter as well as the ground station via the 4G link. These testing showed that our system will able to function in normal circumstances. However, to truly validate that it will work always, we must fly it much further away from the point of launch. The two main issues that prevented us from doing so are the fear of a possibility of a crash, having no further replacements for the UAV and components, as well as the short flight time provided by the battery we have, having a flight time of 10~12 minutes.

The figures below show a sample of the obtained data from the website during flights

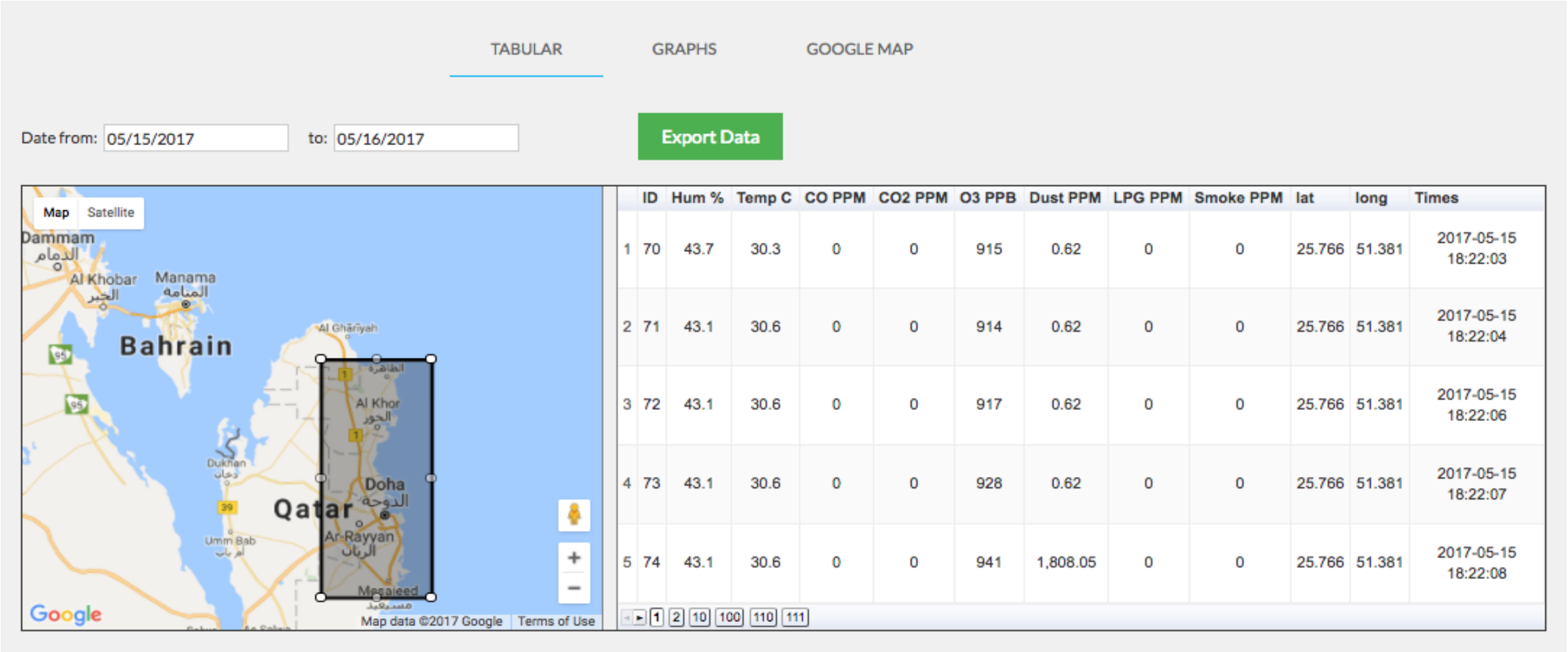

| | ID | Hum % | Temp C | CO PPM | CO2 PPM | O3 PPB | Dust PPM | LPG PPM | Smoke PPM | lat | long | Times |
|---|---|---|---|---|---|---|---|---|---|---|---|---|
| 1 | 70 | 43.7 | 30.3 | 0 | 0 | 915 | 0.62 | 0 | 0 | 25.766 | 51.381 | 2017-05-15 18:22:03 |
| 2 | 71 | 43.1 | 30.6 | 0 | 0 | 914 | 0.62 | 0 | 0 | 25.766 | 51.381 | 2017-05-15 18:22:04 |
| 3 | 72 | 43.1 | 30.6 | 0 | 0 | 917 | 0.62 | 0 | 0 | 25.766 | 51.381 | 2017-05-15 18:22:06 |
| 4 | 73 | 43.1 | 30.6 | 0 | 0 | 928 | 0.62 | 0 | 0 | 25.766 | 51.381 | 2017-05-15 18:22:07 |
| 5 | 74 | 43.1 | 30.6 | 0 | 0 | 941 | 1,808.05 | 0 | 0 | 25.766 | 51.381 | 2017-05-15 18:22:08 |

**Figure 5-17: Sensors reading from the website**

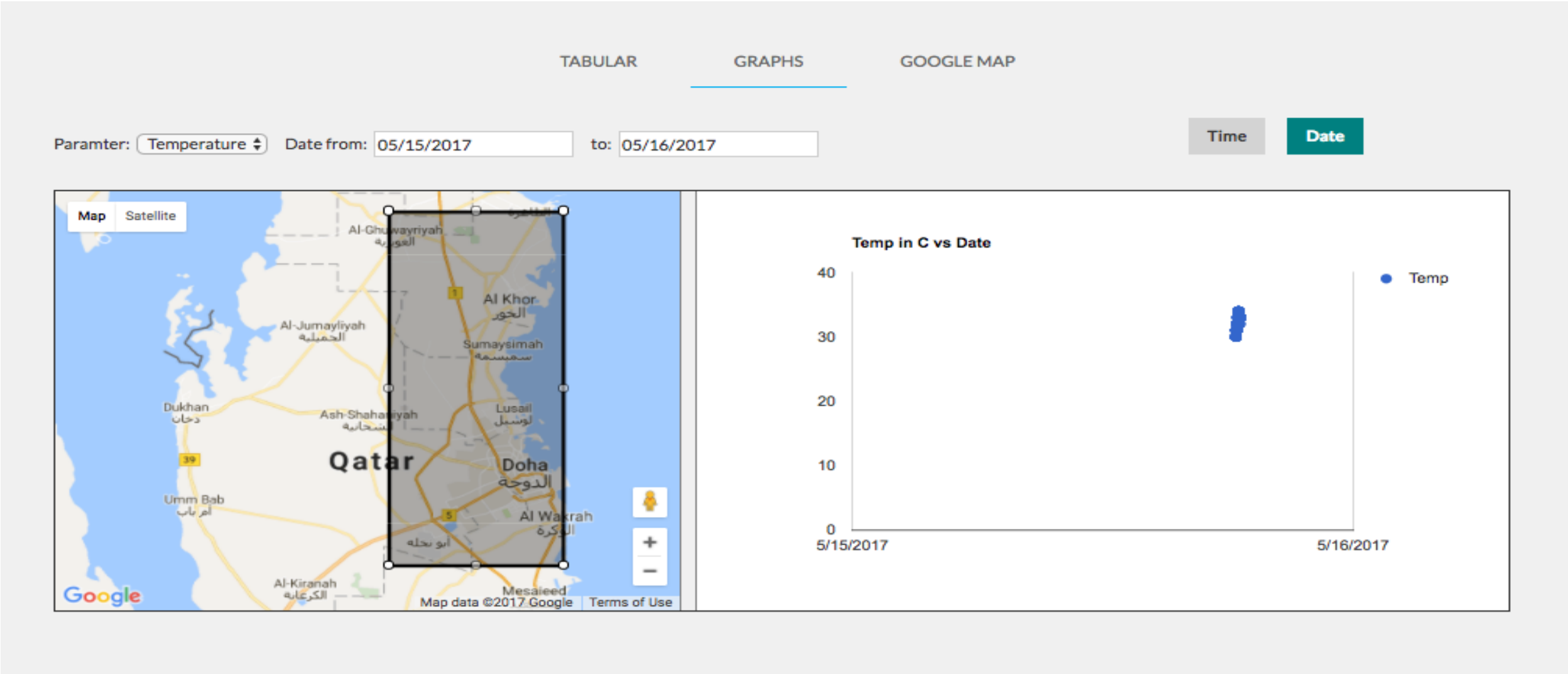

**Figure 5-18: Temperature reading graph by date**

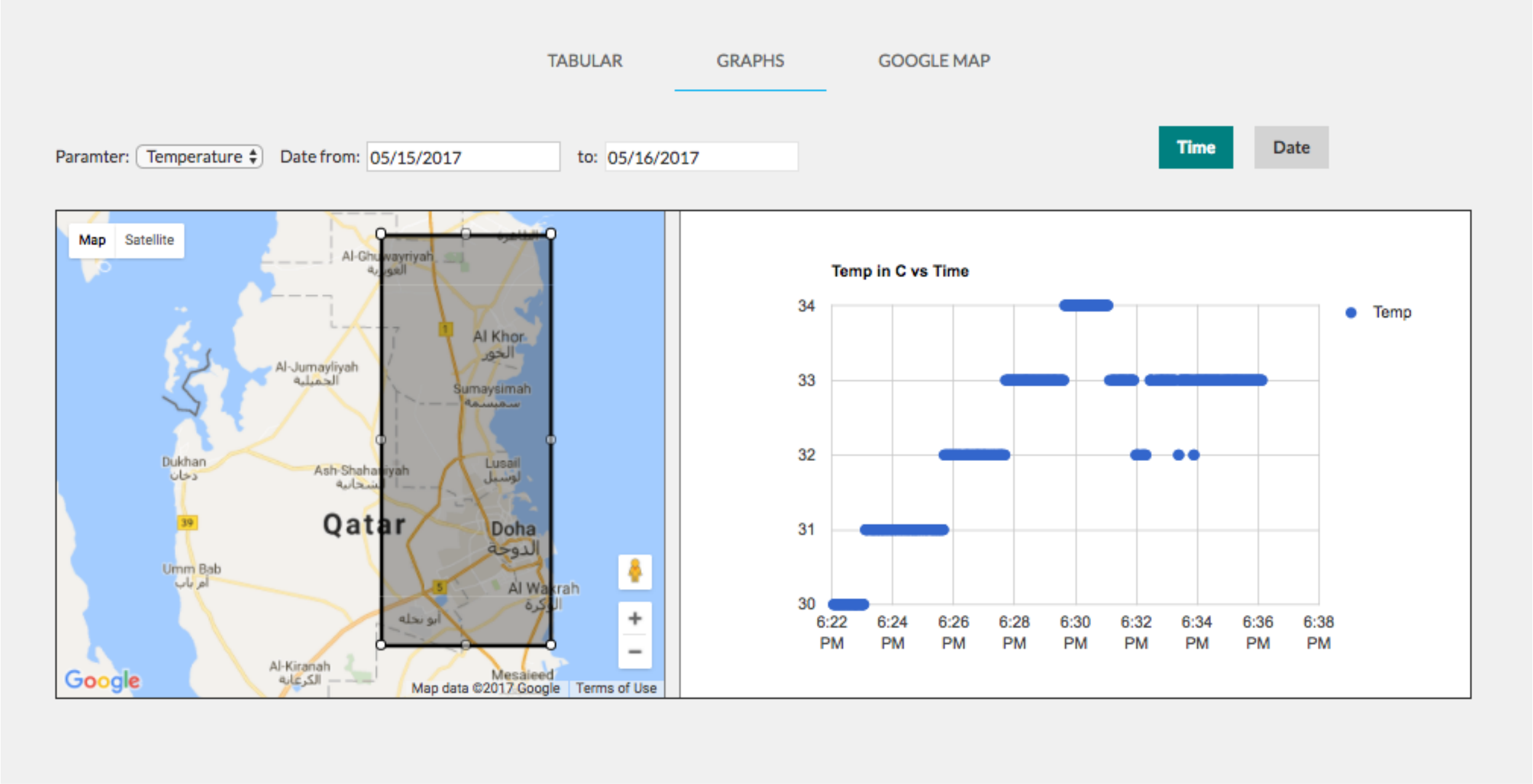

**Figure 5-19:Temperature reading graph by time**

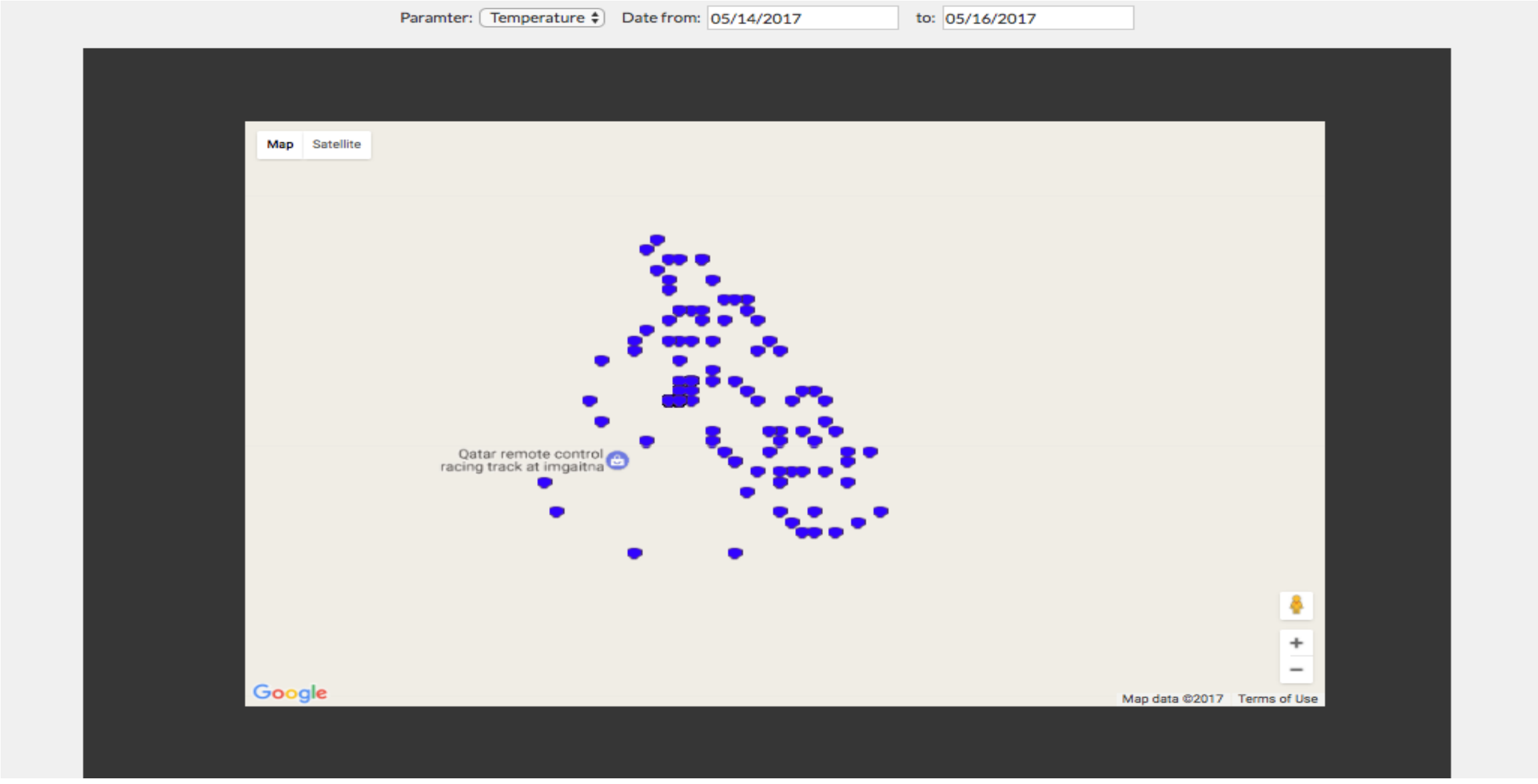

**Figure 5-20: Temperature reading points on Google Map**

# Chapter 6 CONCLUSIONS AND FUTURE WORK

This project aimed to develop RC plane to be autonomous and has specific features such as air quality monitoring and live streaming. The project objective has been successfully achieved where the sensors are successfully interfaced. However, they need proper calibration to give more precise data. The live streaming successfully implemented with a delay of 3~4 seconds, this delay should be further minimized. Moreover, an android application is also successfully implemented to show a map of current measurements as well as the closest recorded measurement. Therefore, the data can be accessed not only through a website but via android application also. Furthermore, autonomous flight was achieved with preset flight paths, to further ease data collection.

There were many challenges in the operation of choosing sensors. In short, the detecting range which will be measured is depending on the mentioned standard is a very accurate range. Obviously, selecting the sensors for the desirable range is very hard, due to the high cost of getting a sensitive output with low cost. Also, a major challenge in implementing UAV applications is due to the limited flying time that is affected by the propeller size; diameter; wing span configuration (rotary instead of fixed wings); payload of the plane and different kinds of the motor. Another main issue was logistics, obtaining a Lithium battery with enough capacity to handle our UAV proved to be tricky, we couldn't find any in Doha or Bahrain due to recent restrictions on Lithium batteries on flights. We ended up buying a battery from Saudi Arabia and bringing it by land. In addition to, the only place we could test our UAV was a RC hobbyist club in the outskirts of Al Khor, which limited our testing to only agreed upon times with members of the club.

After done with the implementation parts there are some parts need to improve to be more efficient in the project. The first thing is calibrating the sensors with calibrating equipment that need a specific equipment. Moreover, increase the flying time of the UAV to have enough to go farther areas and collect data and visualize it in other numerous ways. This could be achieved by utilize a solar system or replace the motor with more efficient one. In addition, building and integrating a mission planner in the website instate of using the local mission controller software. In addition, optimizing the size of the data collection module could be a significant improvement to reduce the weight. Finally, increase the security for the whole system to be safer and reduce the live streaming delay.

# Appendix A SENSORS SPECIFICATIONS

**Table 0-1: SM-PWM-01C (Dust) sensor specification**

<table>
<tr><td>Sensitivity</td><td>The dust sensor can detect 1μm (minimum.)</td></tr>
<tr><td>Sensor response</td><td>This sensor used pulse width signals

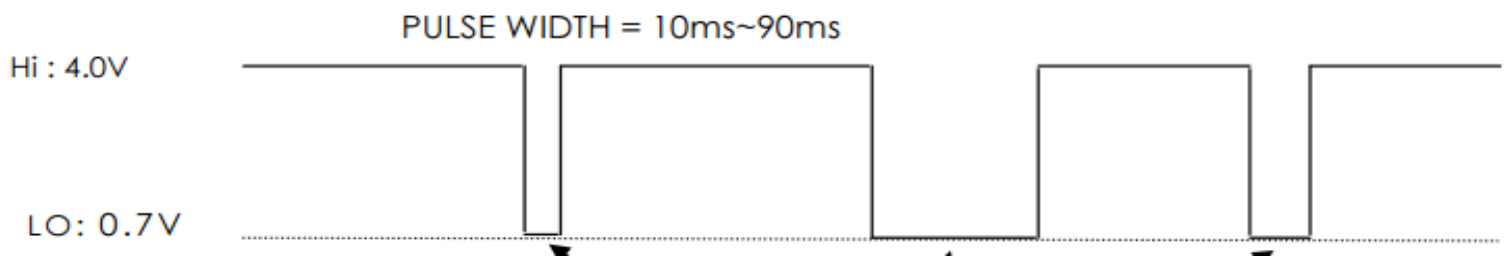

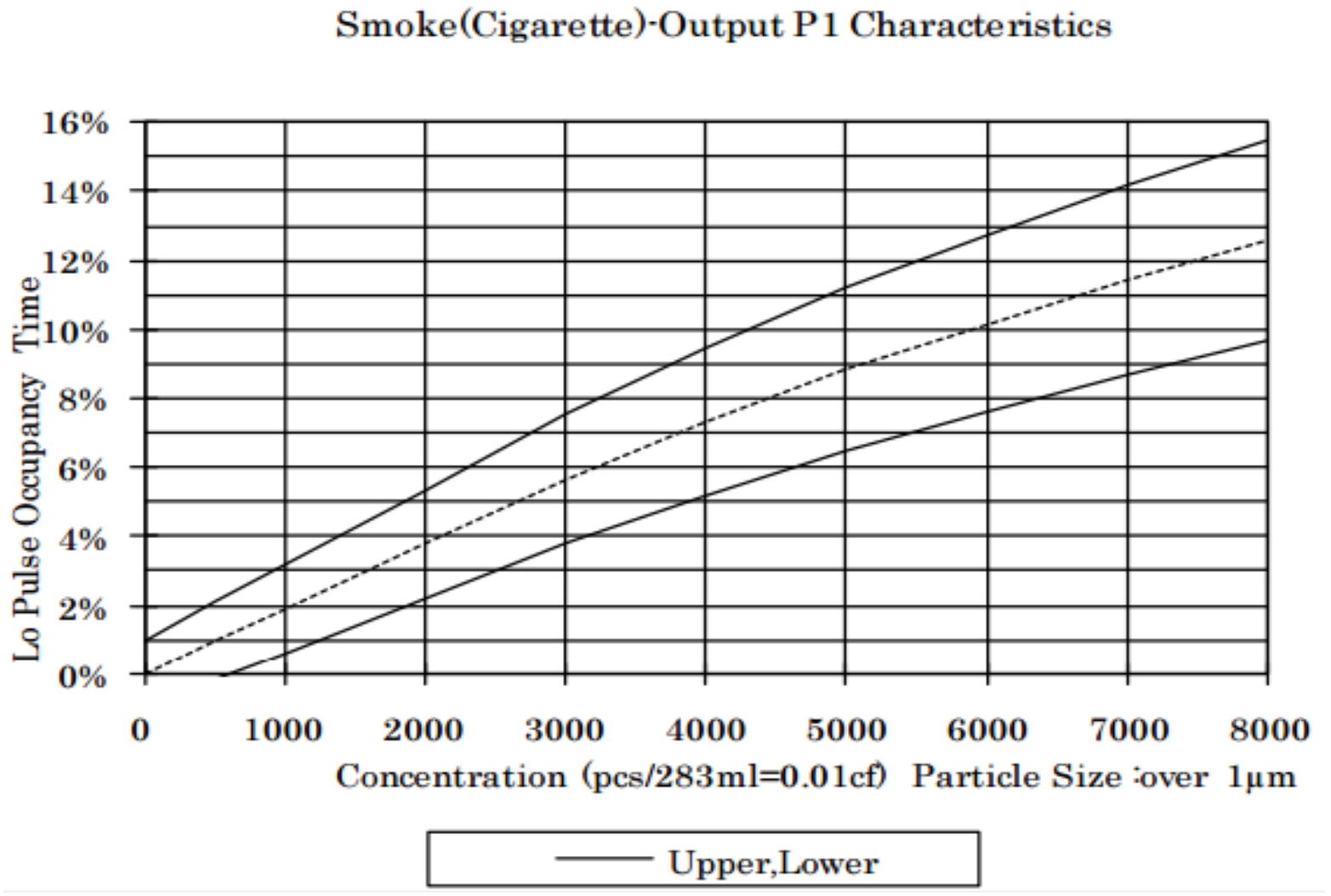

At low pulse 0.7 the reading is not accurate (miss some information) while in the upper voltage it will start from 0</td></tr>
</table>

**Table 0-2: AM2302 (Humidity and temperature) sensor specification**

| | |
|---|---|
| **Sensitivity** | the sensor 25°C at 5V, it does not include hysteresis and nonlinearity perfect results. |
| **Sensor response** | Sensor data bus (SDA) is pulled down to 80μ s, followed by high-80μ s response to host the start signal. 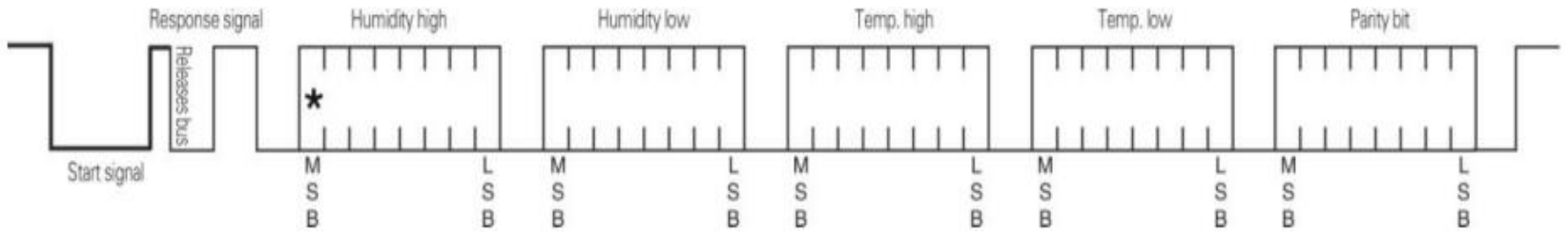  |
| **Temperature effect** | 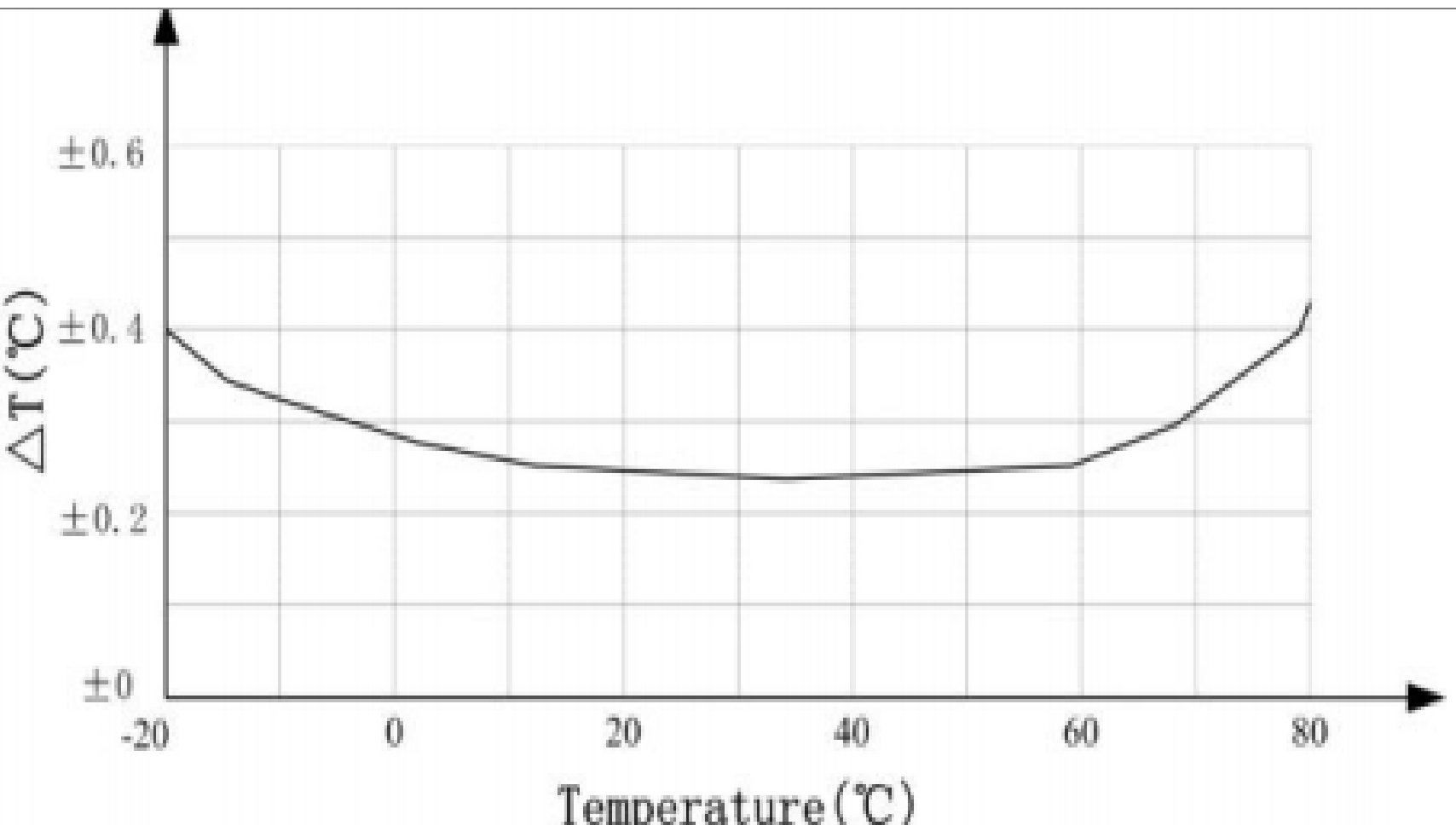  |
| **Humidity relation to temperature change** | 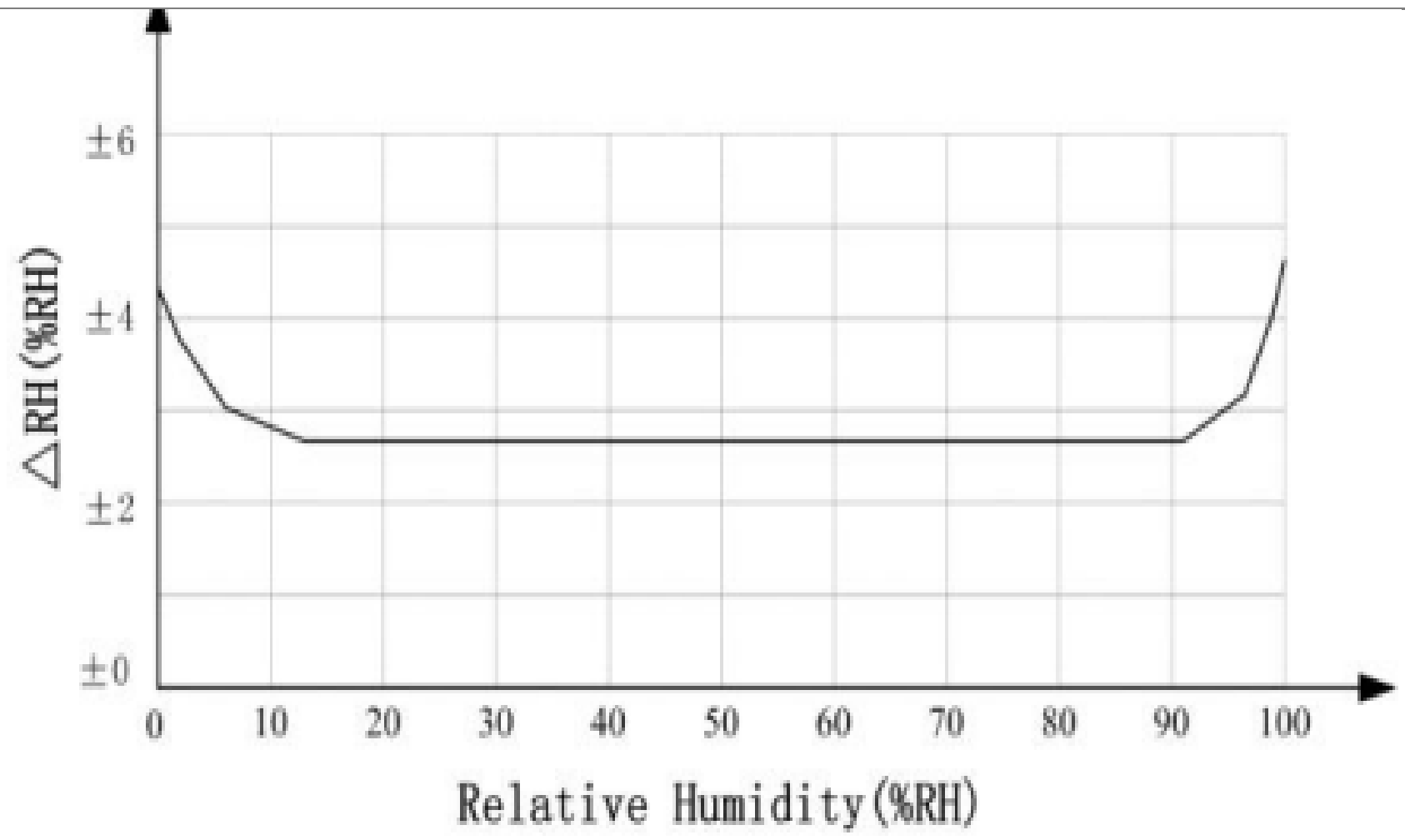  |

# Appendix B SENSORS CODE

```
#include "DHT.h"
#define DHTPIN 2
#define DHTTYPE DHT22
// MQ135
#include "MQ135.h"
#define ANALOGPIN A2    //  Define Analog PIN on Arduino Board CO2
#define RZERO 206.85    //  Define RZERO Calibration Value
///// Dust
int pin = 3;
unsigned long duration;
unsigned long starttime;
unsigned long sampletime_ms = 2000;//sampe 30s ;
unsigned long lowpulseoccupancy = 0;
float ratio = 0;
float concentration = 0;
//// O3
#define         MQ3PIN                      (1)     //define which analog input channel you are going to use
#define         RL3VALUE                     (100)     //define the load resistance on the boar
#define         GAS_O3                      (3)
float           O3Curve[3]  =  {0.69, 0.69, -0.76};
float           Ro3           =  10;
// CO2
MQ135 gasSensor = MQ135(ANALOGPIN);
// CO
#define         MQ_PIN                      (3)     //define which analog input channel you are going to use
#define         RL_VALUE                     (5)     //define the load resistance on the board, in kilo ohms
#define                  RO_CLEAN_AIR_FACTOR                  (9.83) //RO_CLEAR_AIR_FACTOR=(Sensor resistance in clean air)/RO,
#define           CALIBARAION_SAMPLE_TIMES      (50)     //define how many samples you are going to take in the calibration phase
#define           CALIBRATION_SAMPLE_INTERVAL   (500)    //define the time interal(in milisecond) between each samples in the
#define           READ_SAMPLE_INTERVAL          (50)     //define how many samples you are going to take in normal operation
```

```
#define          READ_SAMPLE_TIMES              (5)      //define the time
interal(in milisecond) between each samples in
#define         GAS_LPG                       (0)
#define         GAS_CO                        (1)
#define         GAS_SMOKE                     (2)
float           LPGCurve[3]  =  {2.3, 0.21, -0.47};
float           COCurve[3]  =  {2.3, 0.72, -0.34};
float           SmokeCurve[3] = {2.3, 0.53, -0.44};
float           Ro           =  10;
float sensorValue;
DHT dht(DHTPIN, DHTTYPE);
void setup() {
Serial.begin(9600);
Serial.print("Calibrating...\n");
dht.begin();
pinMode(3, INPUT);
delay(3000);
Ro = MQCalibration(MQ_PIN);                        //Calibrating the sensor.
Please make sure the sensor is in clean air
starttime = millis();//get the current time;
float rzero1 = gasSensor.getRZero();
Ro3 = MQCalibration(MQ3PIN);

}
void loop() {
float h = dht.readHumidity();
float t = dht.readTemperature();
duration = pulseIn(pin, LOW);
lowpulseoccupancy = lowpulseoccupancy + duration;
if ((millis() - starttime) >= sampletime_ms) //if the sampel time = = 30s
{
  ratio = lowpulseoccupancy / (sampletime_ms * 10.0); // Integer percentage
0=&gt;100
  concentration = 1.1 * pow(ratio, 3) - 3.8 * pow(ratio, 2) + 520 * ratio +
0.62; // using spec sheet curve
  lowpulseoccupancy = 0;
  starttime = millis();
}
// O3
sensorValue = MQGetGasPercentage(MQRead(MQ3PIN) / Ro3, GAS_O3);
// Co2
float ppm = gasSensor.getPPM();
```

```
digitalWrite(13, HIGH);

Serial.print(h); // humidity
Serial.print("   ;");
Serial.print(t);// temp
Serial.print("   ;");
Serial.print(concentration); //dust
Serial.print("   ;");
Serial.print(sensorValue);
Serial.print("   ;");
Serial.print(ppm); //co2
Serial.print("   ;");
Serial.print(MQGetGasPercentage(MQRead(MQ_PIN) / Ro, GAS_CO) ); //co
Serial.print("   ;");
Serial.print(MQGetGasPercentage(MQRead(MQ_PIN) / Ro, GAS_LPG) ); //LPG
Serial.print("   ;");
Serial.print(MQGetGasPercentage(MQRead(MQ_PIN) / Ro, GAS_SMOKE) ); //Smoke
Serial.print("   ;");
Serial.print("\n");
}
float MQResistanceCalculation(int raw_adc)
{
return ( ((float)RL_VALUE * (1023 - raw_adc) / raw_adc));
}
float MQCalibration(int mq_pin)
{
int i;
float val = 0;
for (i = 0; i < CALIBARAION_SAMPLE_TIMES; i++) {      //take multiple samples
  val += MQResistanceCalculation(analogRead(mq_pin));
  delay(CALIBRATION_SAMPLE_INTERVAL);
}
val = val / CALIBARAION_SAMPLE_TIMES;                //calculate the average value
val = val / RO_CLEAN_AIR_FACTOR;                     //divided by RO_CLEAN_AIR_FACTOR yields the Ro
//according to the chart in the datasheet
return val;
}
float MQRead(int mq_pin)
{
```

```
int i;
float rs = 0;
for (i = 0; i < READ_SAMPLE_TIMES; i++) {
  rs += MQResistanceCalculation(analogRead(mq_pin));
  delay(READ_SAMPLE_INTERVAL);
}
rs = rs / READ_SAMPLE_TIMES;
return rs;
}
int MQGetGasPercentage(float rs_ro_ratio, int gas_id)
{
if ( gas_id == GAS_LPG ) {
  return MQGetPercentage(rs_ro_ratio, LPGCurve);
} else if ( gas_id == GAS_CO ) {
  return MQGetPercentage(rs_ro_ratio, COCurve);
} else if ( gas_id == GAS_SMOKE ) {
  return MQGetPercentage(rs_ro_ratio, SmokeCurve);
}
else if ( gas_id == GAS_O3 ) {
  return MQGetPercentage(rs_ro_ratio, O3Curve);
}
return 0;
}
int  MQGetPercentage(float rs_ro_ratio, float *pcurve)
{
return (pow(10, ( ((log(rs_ro_ratio) - pcurve[1]) / pcurve[2]) + pcurve[0])));
}
```